\documentclass{article}
\usepackage{microtype}
\usepackage{graphicx}
\usepackage{subcaption}
\usepackage{booktabs} 

\usepackage[dvipsnames]{xcolor}
\usepackage{enumitem}
\usepackage{svg}
\usepackage{bm}
\usepackage{soul}
\usepackage{setspace}
\usepackage{sidecap}

\usepackage{hyperref}

\usepackage{tikz}
\usepackage{pgfplots}
\usetikzlibrary{positioning, fit, shapes.geometric, bending, decorations.text, matrix, intersections, pgfplots.groupplots}
\usepgfplotslibrary{fillbetween}
\pgfplotsset{compat=1.18} 

\definecolor{C0}{HTML}{3064B8}
\definecolor{C1}{HTML}{9F6ABF}
\definecolor{C2}{HTML}{D94848}
\definecolor{C3}{HTML}{E88E3B}

\pgfplotscreateplotcyclelist{mygradient}{
{C0}, {C1}, {C2}, {C3}
}

\pgfplotsset{
try min ticks=3,
legend image code/.code={
\draw[mark repeat=2,mark phase=2]
plot coordinates {
(0cm,0cm)
(0.15cm,0cm)        
(0.3cm,0cm)         
};%
}
}

\PassOptionsToPackage{numbers, sort&compress}{natbib}

\usepackage[preprint]{neurips_2026}

\setcitestyle{authoryear,round} 

\usepackage[T1]{fontenc}
\usepackage{url}
\usepackage{amsfonts}
\usepackage{nicefrac}

\usepackage{amsmath}
\usepackage{amssymb}
\usepackage{mathtools}
\usepackage{amsthm}
\usepackage{xcolor}

\hypersetup{
    colorlinks,
    linkcolor={blue!64!black},
    citecolor={blue!64!black},
    urlcolor={blue!64!black},
    pdfborder={0 0 0}
}

\usepackage[capitalize,noabbrev]{cleveref}

\theoremstyle{plain}
\newtheorem{theorem}{Theorem}[section]
\newtheorem{proposition}[theorem]{Proposition}
\newtheorem{lemma}[theorem]{Lemma}

\theoremstyle{definition}
\newtheorem{definition}[theorem]{Definition}
\newtheorem{assumption}[theorem]{Assumption}
\theoremstyle{remark}
\newtheorem{remark}[theorem]{Remark}

\newcommand{\bct}{\bm{c}(t)}
\newcommand{\bD}{\bm{D}}
\newcommand{\bDdag}{\bD^{\dagger}}
\newcommand{\bI}{\bm{I}}
\newcommand{\bmu}{\bm{\mu}}
\newcommand{\br}{\bm{r} }
\newcommand{\brt}{\br(t)}
\newcommand{\bSigma}{\bm{\Sigma}}
\newcommand{\bs}{\bm{s}}
\newcommand{\bst}{\bs(t)}
\newcommand{\bvt}{\bm{v}(t)}
\newcommand{\but}{\bm{u}(t)}

\newcommand{\dt}{\Delta t}
\newcommand{\drt}[1]{\Delta \br_{#1}(t)}
\newcommand{\En}[1]{\mathbb{E}_{\bm{\eta}} \norm{#1}}
\newcommand{\friction}{\lambda_v}

\newcommand{\loss}{\mathcal{L}(t)}

\newcommand{\norm}[1]{\lVert #1 \rVert_2}
\newcommand{\score}{\nabla_{\brt} \log p(\brt)}
\newcommand{\sigmaeta}{\sigma_r \bm{\eta}(t)}
\newcommand{\sigmadt}{\sigma_r^2 \dt}
\newcommand{\timeset}[1]{ \left\{ #1 \right\}_{t=0}^T }

\DeclareMathOperator*{\argmin}{arg\,min}

\newcommand{\column}[1]{\begin{tabular}{c} #1 \end{tabular}}

\usepackage[textsize=tiny]{todonotes}

\setlist[enumerate]{leftmargin=5mm} 
\setlist[itemize]{leftmargin=5mm}

\begin{document}

\title{Second-Order Dynamics in Noisy RNNs Produce Exploration and Temporal Compression in Replay}

\author{%
Josue Casco-Rodriguez \\
Department of Electrical and Computer Engineering \\
Rice University, USA \\
\texttt{jc135@rice.edu} \\
\And
Nanda H. Krishna \\
Mila - Quebec AI Institute \\
\& Universit\'{e} de Montr\'{e}al, Canada \\
\And
Richard G. Baraniuk \\
Department of Electrical and Computer Engineering \\
  Rice University, USA \\
}

\maketitle

\begin{abstract}
Biological neural networks (like the hippocampus) can internally generate ``replay'' resembling stimulus-driven activity.
Recent computational models of replay use noisy recurrent neural networks (RNNs) trained to path-integrate.
Replay in these networks is described as Langevin sampling, but this theory alone fails to capture the environmental exploration and temporal compression observed in real biological replay.
To close this gap, we re-examine spontaneous noisy RNN activity (``replay'') as Langevin sampling, one timestep at a time, from a time-series distribution resembling lived experiences (unlike previous works which defined sampling as Monte Carlo inference over a static distribution).
We provide a normative second-order Langevin sampling framework wherein spike-frequency adaptation drives spatial exploration at the cost of speed, and momentum via dendritic input differentiation drives temporal compression.
We numerically verify our framework via simulations of noisy RNNs trained on 2D triangular and T-maze navigation activity, and others trained on place cell activity of synthetic rats.
Crucially, we observe that simultaneous adaptation and momentum increases both replay exploration and speed, and provide one of the first accounts for biologically plausible temporal compression in rate-based RNN replay.
Our findings encourage experiments into whether dendritic input differentiation, or other mechanisms, may support temporal compression in memory circuits like the hippocampus.
\end{abstract}

\vspace{-1em}
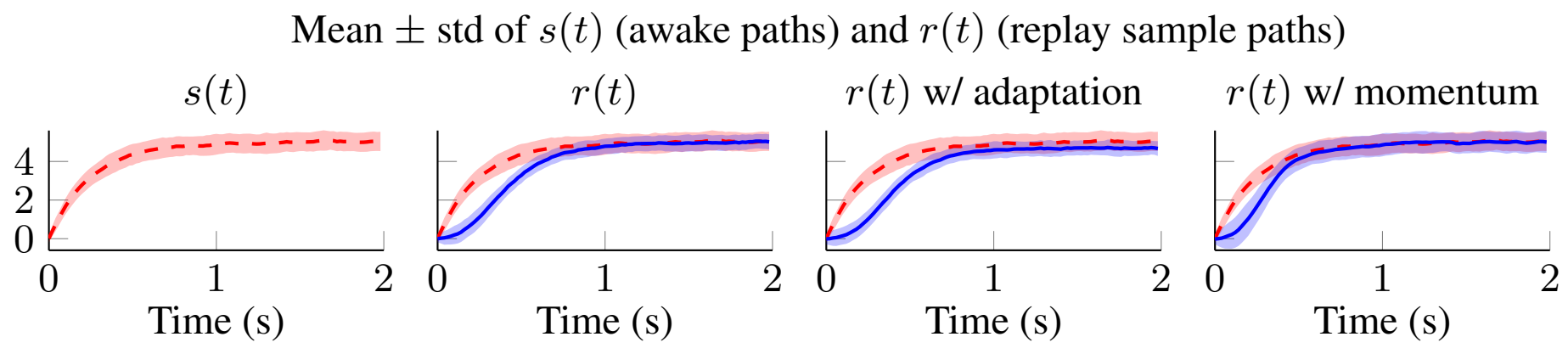
\begin{figure}[h]
\centering
\begin{tikzpicture}

\small 

\newcommand{\plotOU}[3]{
    \addplot[#1, #2, thick] table[x=time, y=mean_#3] {means/1d_ou.txt};
    \addplot[name path=upper, draw=none] table[x=time, y expr = \thisrow{mean_#3} + \thisrow{std_#3}] {means/1d_ou.txt};
    \addplot[name path=lower, draw=none] table[x=time, y expr = \thisrow{mean_#3} - \thisrow{std_#3}] {means/1d_ou.txt};
    \addplot[#1, nearly transparent] fill between[of=upper and lower];
}

\node at (6, 1.7) {Mean $\pm$ std of $s(t)$ (awake paths) and $r(t)$ (replay sample paths)};

\begin{groupplot}[
    group style = {
        group size = 4 by 1,
        horizontal sep = 0.03\linewidth},
    width = 0.3\linewidth,
    height = 0.18\linewidth,
    title style={yshift=-2mm},
    xlabel=Time (s),
    xlabel shift={-1mm},
    xmin=0, xmax=2,
    ymin=-0.6, ymax=5.6,
    xtick={0, 1, 2},
    axis x line*=bottom,
    axis y line*=left,
    table/col sep=comma
    ]

\nextgroupplot[title=$s(t)$]
    \plotOU{red}{dashed}{s};

\nextgroupplot[title=$r(t)$, yticklabels={}]
    \plotOU{red}{dashed}{s};
    \plotOU{blue}{}{overdamped};

\nextgroupplot[title= $r(t)$ w/ adaptation, yticklabels={}]
    \plotOU{red}{dashed}{s};
    \plotOU{blue}{}{adaptation};

\nextgroupplot[title= $r(t)$ w/ momentum, yticklabels={}]
    \plotOU{red}{dashed}{s};
    \plotOU{blue}{}{underdamped};

\end{groupplot}
\end{tikzpicture}
\caption{
\textbf{Momentum accelerates offline replay, adaptation slows it.}
Here we simulate a noisy RNN $r(t)$ that optimally path-integrates an Ornstein-Uhlenbeck process $s(t)$ from its velocity $s'(t)$.
We assume $r(t)$ minimizes the loss in \Cref{eq:loss_upper} and thus evolves according to its \textit{score function} $\nabla_{r(t)} \log p(r(t))$ (Equations \ref{eq:rnn_langevin} and \ref{eq:score_ou_main}), performing a variant of Langevin sampling when no input is given.
Above, we compare three modifiers of RNN activity: no modifiers (a.k.a. \textit{overdamped}), spike-frequency \textit{adaptation} (negative feedback), and our proposed \textit{momentum} (which directly implements \textit{underdamped} dynamics).
Each modifier affects the replay time-series distribution $p(r(t))$ in different ways:
adaptation slows $p(r(t))$ and pushes it away from its natural steady states, while momentum accelerates the convergence of $p(r(t))$ towards its steady states.
}
\label{fig:1d_mean_std}
\end{figure}
\vspace{-0.5em}
\section{Introduction}

During quiescent periods such as sleep or wakeful resting, some neural circuits internally generate activity resembling that of active periods \citep{tingley2020methods}. Such ``replay'' phenomena have been observed in the prefrontal \citep{euston2007fast, peyrache2009replay}, sensory \citep{kenet2003spontaneously, xu2012activity}, motor \citep{hoffman2002coordinated}, and entorhinal cortices \citep{gardner2022toroidal}; the anterior thalamus \citep{peyrache2015internally}; and the hippocampus \citep{hippocampal_sharp_waves, replay_of_neuronal, nadasdy1999replay, lee2002memory, foster2017replay}.
Of these circuits, the hippocampus is particularly interesting because its robustness in tasks like navigation \citep{hippocampus_cognitive_map, model_hippocampal, deciphering_hippocampal} and planning \citep{hippocampal_place-cell_sequences, dorsal_hippocampus} during active states seems crucially tied to its spontaneous activity during quiescent states \citep{two-stage_model, hippocampal_sharp_wave_ripple, sleep_price, olafsdottir2015hippocampal,olafsdttir2018therole}.

While some works have produced replay using supervised generative models \citep{deperrois2022learning}, most existing models of hippocampal activity treat replay as an emergent byproduct of careful network design. Relevant network parameters include connectivity structures \citep{shen1996modeling, milstein2023offline}, local plasticity mechanisms \citep{hopfield2009neurodynamics, litwin2014formation, theodoni2018theta, haga2018recurrent, predictive_learning_rules}, firing rate adaptation \citep{chu2023firing, computational_model_preplay, cell_assembly_sequences, noisy_adaptation, dynamics_of_adaptive_canns}, and input modulation \citep{kang2019replay}. While these models reproduce aspects of replay, they are typically motivated by empirical findings and lack theoretical justification. 

A more principled model of hippocampal function with emergent replay is path-integration \citep{sufficient_conditions, sequential_predictive_learning}, wherein neural circuits predict dynamic environment or task variables from imperfect observations thereof, e.g., tracking position from velocity measurements.
This normative description of the hippocampus as a sequence predictor \citep{computational_approach,hippocampus_predictive_map} matches hippocampal encodings of upcoming stimuli \citep{hippocampus_preserves_order} and prediction errors \citep{hippocampal_representations_switch, emergence_predictive_model}, and neural activity sweeps that represent possible future trajectories \citep{constant_sub-second, neural_ensembles_ca3}.
Unlike traditional, hand-crafted models of hippocampal circuits (i.e., continuous attractor networks \citep{path_integration, attractor_neural_networks_storing}), sequential predictive learning models are trained from data.
Nonetheless, they account for the emergence of place cells \citep{predictive_learning_network_mechanism, sequential_predictive_learning, predictive_sequence_learning}, grid cells \citep{cueva2018emergence, sorscher2019unified}, and head direction cells \citep{Cueva2020Emergence, Uria2020Spatial}; can incorporate phenomena like theta oscillations \citep{sequential_predictive_learning}; and exhibit quiescent replay activity \citep{sufficient_conditions, sequential_predictive_learning, predictive_sequence_learning}.

\begin{figure}[h]
\centering
\begin{tikzpicture}

\node[inner sep=0pt] (img) {\includegraphics[
trim={0mm 2cm 2cm 0mm}, clip, width=0.35\linewidth
]{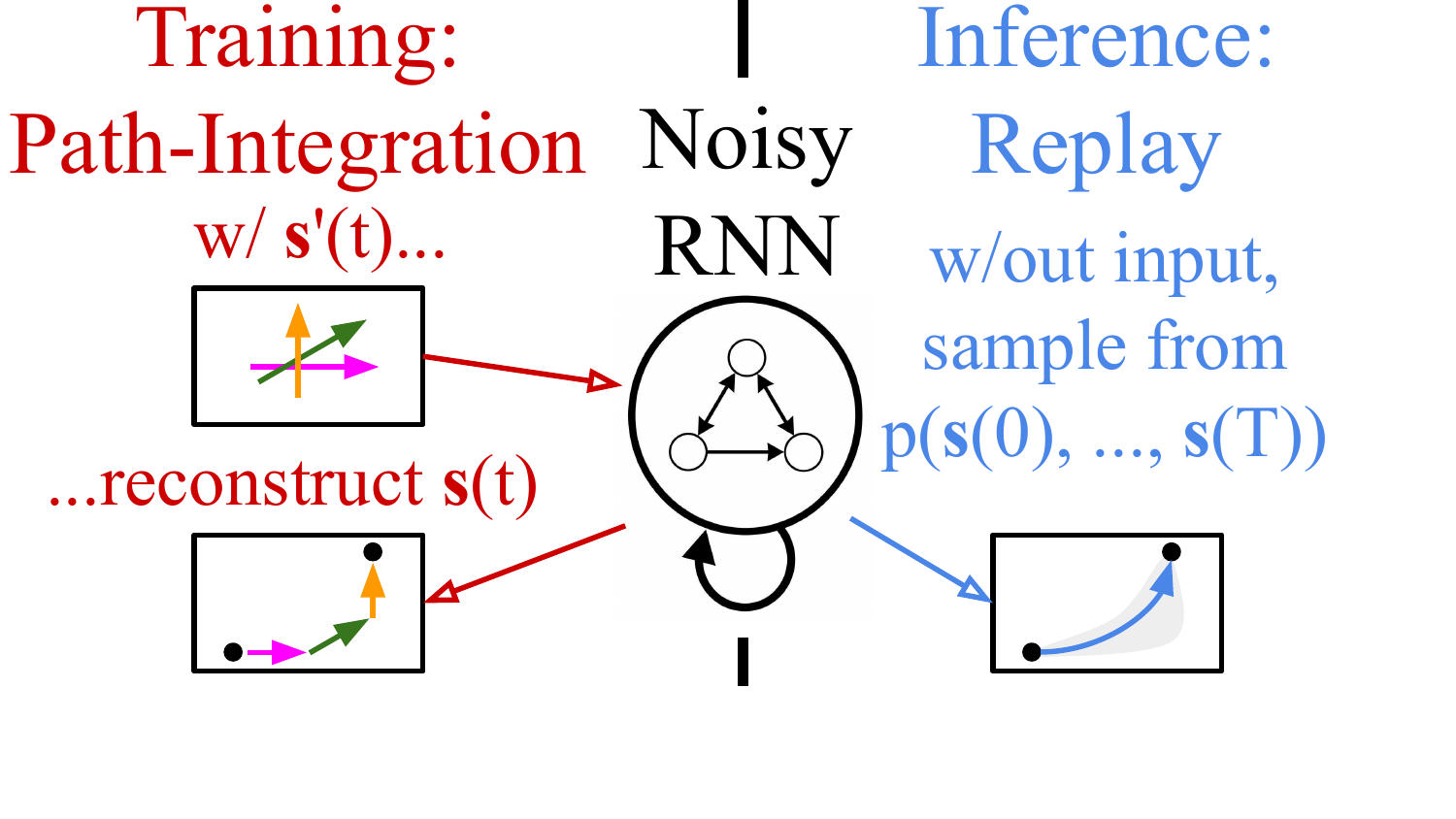}};

\node[inner sep=0pt, right=3mm of img.east, anchor=west] (T1) {\includegraphics[
trim={0mm 9mm 0mm 9mm}, clip, width=0.32\linewidth
]{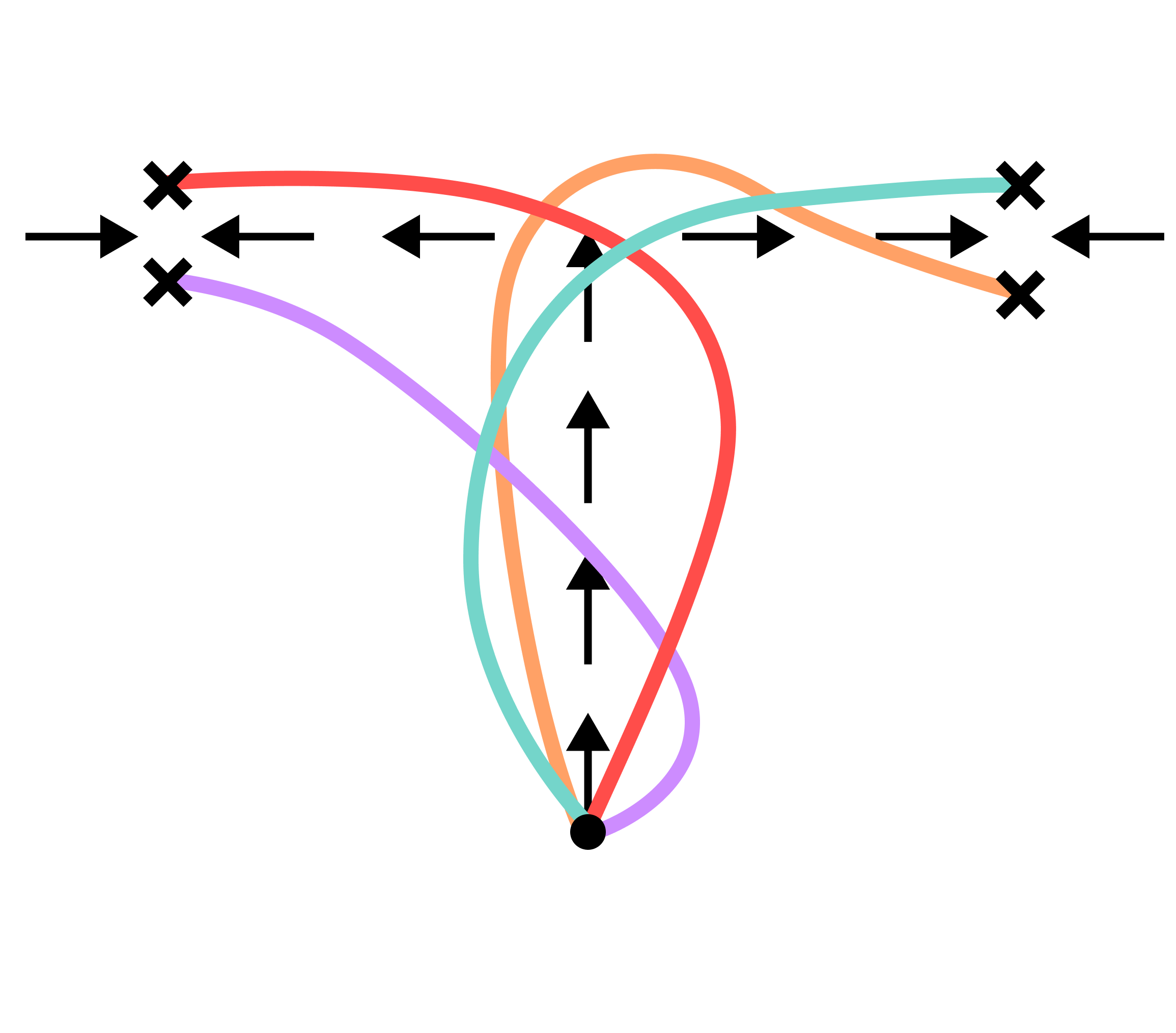}};

\node[inner sep=0pt, right=3mm of T1.east, anchor=west] (T2) {\includegraphics[
trim={0mm 9mm 0mm 9mm}, clip, width=0.3\linewidth
]{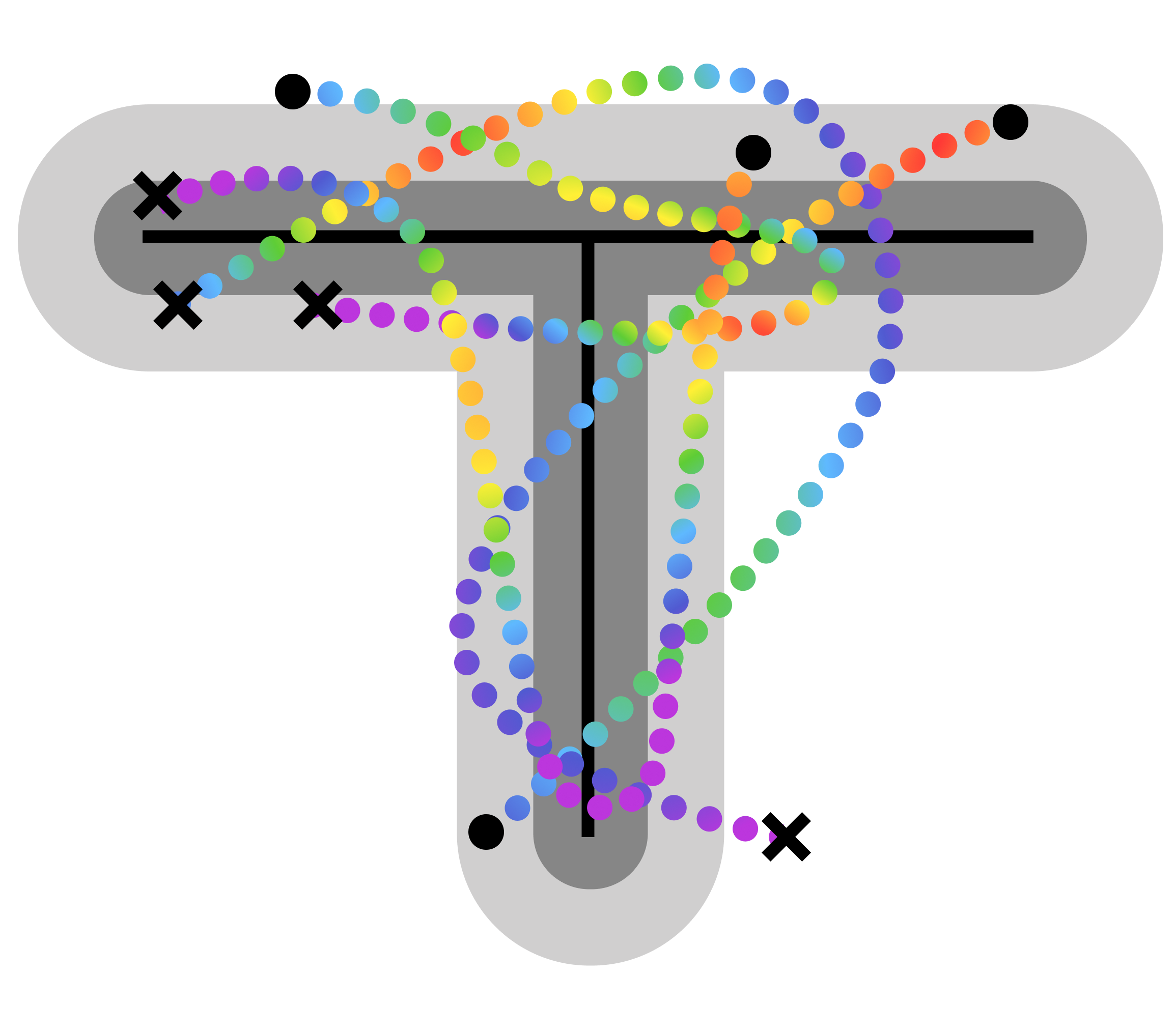}};

\node[inner sep=0pt] (bar) at (7,-1) {\includegraphics[width=0.15\linewidth, height=3mm]{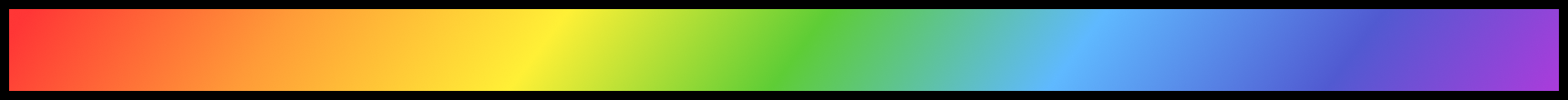}};
\node[inner sep=0pt, above=0pt of bar.north] {\small \column{
different color =
\\
different sample
}};

\node[anchor=south, inner sep=0pt, above=3mm of img.north] {\small \column{
\textbf{(a) Path-integration yields}
\\
\textbf{generative replay}
}};

\node[anchor=south, inner sep=0pt, above=-1mm of T1.north] {\small \column{
\textbf{(b) Internally-driven replay of}
\\
\textbf{directed time-series} 
}};

\node[anchor=south, inner sep=0pt, above=0mm of T2.north] {\small \column{
\textbf{(c) Input-driven Monte Carlo}
\\
\textbf{sampling through time} 
}};

\end{tikzpicture}
\caption{
\textbf{Noisy RNNs trained to path-integrate can internally generate replay trajectories resembling lived experiences.}
\textbf{(a)} A noisy RNN learns to \textcolor{red}{path-integrate}: it estimates a signal $\bst$ from (nonlinear) observations of velocity $\bm{s}'(t)$.
Without input, this RNN's noise-driven activity $\brt$ samples from the time-series distribution of awake activity, one timestep at a time (“\textcolor{blue}{replay}”).
\textbf{(b)} Each internally-driven replay trajectory is a single sample from a time-series distribution with strict temporal dependencies: awake trajectories start and stop at certain positions in specific orders, and replay trajectories reproduce these temporal structures.
\textbf{(c)} Our definition of replay as sampling is distinct from other definitions of neural sampling in which stochastic input-driven trajectories are used to gather samples to perform Monte Carlo inference on a static distribution without temporal dependencies (see \Cref{sec:replay_vs_other_sampling}).
}
\label{fig:setup_overview}
\end{figure}
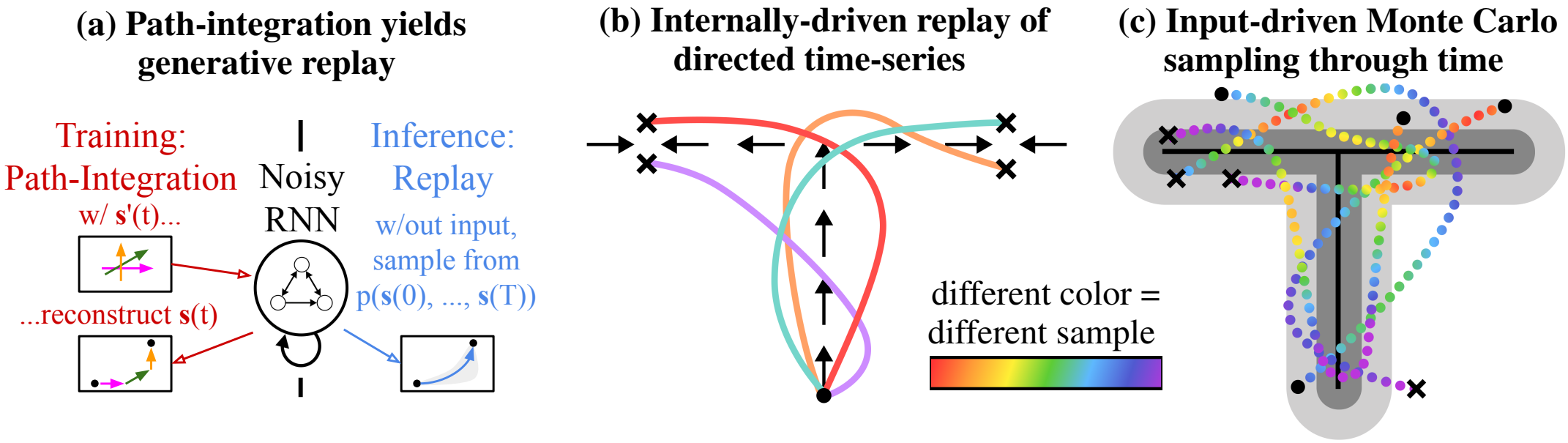

\citet{sufficient_conditions} provided the first theoretical foundation for replay in noisy path-integrating networks, proving their quiescent (i.e., noise-driven, without inputs) activity is diffusive replay \citep{hippocampal_reactivation_random} generated via Langevin sampling \citep{besag1994}, one timestep at a time, from the time-series distribution of awake activity.
Crucially, this definition of time-series sampling for replay is different from how modern AI models generate time-series---RNN replay is made one timestep at a time, without modifying any other timesteps, whereas AI models modify all timesteps simultaneously---and different from other definitions of sampling in computational neuroscience---replay samples from a temporally structured time-series distribution resembling lived experiences, whereas other types of sampling  \citep{adaptation_accelerating, hamiltonian_brain, fast_sampling-based, natural_gradient_bayesian} use external inputs to generate temporally unstructured time-series whose individual timesteps are gathered for Monte Carlo calculations (\Cref{fig:setup_overview}).

Afterwards, \citet{sequential_predictive_learning} introduced a mechanism to induce \textit{exploration} in replay via spike-frequency adaptation, i.e., negative feedback. Exploration is the notion that replay expresses a variety of behavioral sequences \citep{hippocampal_replay_extended,content_of_hippocampal}, and is associated with long trajectories in neural space, visitation of multiple attractor basins, and transitions that were not present in awake activity. Adaptation can destabilize attractors and induce sudden transitions in replay activity \citep{cell_assembly_sequences,noisy_adaptation,dynamics_of_adaptive_canns,sequential_predictive_learning}, thereby facilitating exploration, and is thought to play a key role in replay dynamics \textit{in vivo} \citep{nrem_sleep}. However, existing theory on noisy path-integrating networks does not account for these mechanisms, or capture the \textit{temporal compression} of replay sequences relative to awake sequences of activity \textit{in vivo} \citep{nadasdy1999replay, hippocampal_sharp_wave_ripple, speed_time_compressed, emergence_preconfigured}.
To our knowledge, temporal compression is not captured in replay from any trained RNN model, and so far our understanding of its mechanisms is limited to short-term facilitation and phase precession in hand-crafted spiking networks \citep{temporal_compression_mediated,phase_precession_neural_code}.

Overall, while noisy path-integration is a promising model of hippocampal function and replay, its theory leaves crucial details of biological replay unaccounted for: exploration and temporal compression.
We close this gap with a normative second-order Langevin sampling framework for spontaneous neural activity (replay).
We show that spike-frequency adaptation drives spatial exploration but introduces a dynamical drag that slows replay.
To recover replay speeds more akin to biology, we introduce momentum via dendritic input differentiation, and relate it to underdamped Langevin sampling.
Using noisy RNNs in several continuous navigation tasks, we demonstrate that introducing both adaptation and momentum in spontaneous RNN dynamics enables replay trajectories to visit multiple attractor basins (exploration) while moving quickly (temporal compression).
These results provide the first formal account of temporal compression in rate-based predictive networks and offer testable predictions regarding temporal compression and dendritic processing in memory circuits like the hippocampus.

\section{Background: Replay as Langevin Sampling}

Neural circuits in several regions of the brain, particularly memory areas like the hippocampus, generate spontaneous replay of activity from when the organism was active or awake.
Traditional models have replay emerge from hand-crafted networks, but we are interested in recent models that prove replay emerges as a byproduct of learning in path-integrating networks like the hippocampus \citep{sufficient_conditions}.
Without external inputs (e.g., sensory cues), these networks undergo Langevin dynamics to sample from the distribution of awake activity they were trained on.
Here we: (1) define Langevin dynamics, (2) summarize how path-integrating noisy RNNs implicitly learn generative Langevin dynamics, (3) specify the crucial distinction between sampling in replay and sampling in other fields of computational neuroscience, and (4) establish the limitations of first-order Langevin sampling that motivate second-order dynamics for replay in neural circuits.
In \Cref{appendix:deviations_langevin} we discuss the exact nuances in characterizing non-stationary noisy RNN activity (replay) as Langevin sampling.

\subsection{Classical Langevin Dynamics}

Here  we define Langevin dynamics, without (Equation \ref{eq:overdamped_langevin}) and with (Equations \ref{eq:underdamped_langevin_2nd_order} and \ref{eq:underdamped_langevin_1st_order}) momentum, as an iterative process by which an agent could attempt to sample from an unknown distribution $p(x)$ using only its \textit{score function} $\nabla \log p(x)$.
%
Langevin sampling from $p(x)$ is stochastic gradient ascent of an iterate $x(t)$ up the log-likelihood of $p(x)$ via its score function:
\begin{equation} \label{eq:overdamped_langevin}
    \dot{x}(t) = \nabla \log p(x) + \sqrt{2} \eta(t),
\end{equation}
where $\eta(t)$ is Gaussian white noise. While Equation \ref{eq:overdamped_langevin} describes \textit{overdamped} dynamics, there are also \textit{underdamped} (i.e., momentum) Langevin dynamics\footnote{Discretized underdamped Langevin dynamics are related to Hamiltonian MCMC \citep{underdamped_langevin_mcmc}.} (Equations \ref{eq:underdamped_langevin_2nd_order} or \ref{eq:underdamped_langevin_1st_order}, see Chapter 6 of \citet{stochastic_processes}) that converge faster to the target distribution $p(x)$ and better utilize noisy gradients \citep{underdamped_langevin_mcmc}:
\begin{gather} 
    \ddot{x}(t) = \nabla \log p(x) - \gamma \dot{x}(t) + \sqrt{2 \gamma} \eta(t), \quad \mathrm{or}
    \label{eq:underdamped_langevin_2nd_order}
    \\
    \dot{x}(t) = v(t), \enspace \dot{v}(t) = \nabla \log p(x) - \gamma v(t) + \sqrt{2 \gamma} \eta(t)
    \label{eq:underdamped_langevin_1st_order}
\end{gather}


\subsection{Replay in Path-Integrating Noisy RNNs}

Biological neural circuits like the hippocampus can navigate through environments when awake, and recall navigatory episodes during rest. Accordingly, we summarize the proof from \citet{sufficient_conditions} that a noisy RNN trained on a biologically relevant navigation task like path-integration can act as a generative model, as illustrated in \Cref{fig:setup_overview}.

Definitions \ref{definition:rnn} and \ref{definition:path-integration} express that our RNNs have nonlinear dynamics and internal noise, and learn to estimate a signal $\bst$ by integrating observations of $\boldsymbol{s}'(t)$. This ability, known as path-integration, underlies navigation and the estimation of positions from motion cues \citep{neural_dynamics_landmark, human_path_integration}.

\begin{definition}
\label{definition:rnn}
A \emph{noisy recurrent neural network (RNN)} has neural activity $\brt$ that evolves at each timestep $t$ via a learnable nonlinear function $f(\cdot)$ of its previous hidden states, an input signal $\but$, and white noise $\bm{\eta}(t)$; unless otherwise stated (see \Cref{appendix:masked_training}), $\bm{u}(t) = \bs'(t)$ or a nonlinear observation thereof:
\begin{equation}
    \label{eq:rnn_definition}
    \br(t + \dt) = f(\brt, \but, \sigmaeta), \quad \sigmaeta \sim \mathcal{N}(0, \bm{I} \sigma_r^2 \dt)
    \color{black}
\end{equation}
\end{definition}

\begin{definition}
\label{definition:path-integration}
A \textit{path-integration} objective $\loss{}$ penalizes the difference between a state variable $\bst$ and a learnable linear projection of the RNN activity $\brt$:
\begin{equation}
    \loss = \En{\bst - \bD\brt}
\end{equation}
\end{definition}

Assumption \ref{assumption:gaussian_r(t)} is that the optimal RNN dynamics accurately path-integrate, such that the conditional probability distribution of network activity given the environmental signal (e.g., position) is Gaussian, and all variability comes from the intrinsic noise in the RNN dynamics. 
Assumption \ref{assumption:additive_decomposition} is that RNN dynamics decompose into additive components.
Finally, Assumption \ref{assumption:greedy} is that the RNN dynamics are greedily optimal for path-integration---the network does the best that it can at each timestep, assuming at each previous timestep the best possible job has been done.

\begin{assumption}
\label{assumption:gaussian_r(t)}
The optimal $\brt$ minimizes $\loss$ such that $p(\br^*(t))$ is normal around $\bm{D}^\dagger \bst$:
\begin{equation}
    p(\br^*(t) | \bst) \sim \mathcal{N}(\bDdag \bst, \, \bI \sigmadt)
    \label{eq:p_r|s}
\end{equation}
\end{assumption}

\begin{assumption}
\label{assumption:additive_decomposition}
Noisy RNN dynamics decompose into two
functions\footnote{
Each function $\drt{1}, \drt{2}$ can depend on
variables beyond $t$, but they are omitted for concision.
} and white noise:
\begin{equation}
\label{eq:rnn_update_definition}
    \drt{} = \br(t + \dt) - \brt 
    \approx \drt{1} + \drt{2} + \sigmaeta,
    \quad
    \sigmaeta \sim \mathcal{N}(0, \sigma_r^2 \dt)
\end{equation}
\end{assumption}

\begin{lemma}
\label{lemma:upper_bound}
With Assumptions \ref{assumption:additive_decomposition} and \ref{assumption:gaussian_r(t)},  $\mathcal{L}(t+\Delta t)$ is upper bounded by $\widehat{\mathcal{L}}(t+\Delta t)$:
\begin{equation}
    \label{eq:loss_upper}
    \widehat{\mathcal{L}}(t+\Delta t) = \lVert \bs'(t) - \bD \drt{2} \rVert + \En{\bm{D}\sigma_r\boldsymbol{\eta}(t)}
    + \lVert \bm{D} \rVert_F \En{ \drt{1} + \sigma_r\boldsymbol{\eta}(t-\Delta t)}
\end{equation}
\end{lemma}

\begin{assumption}
\label{assumption:greedy}
The optimal $\brt$ greedily minimizes $\widehat{\mathcal{L}}$ at $t$ only ($T$ is the final timestep):
\begin{equation}
    \timeset{\br^*(t)} = \argmin_{\timeset{\brt}} \sum_{t=0}^T \mathcal{L}(t+\Delta t) \Delta t
    \approx \biggl\{ \argmin_{\brt} \widehat{\mathcal{L}}(t+\Delta t) \biggr\}_{t=0}^T
\end{equation}
\end{assumption}

Theorem \ref{theorem:optimal_update} states that path-integrating RNNs will use their learned score function of awake activity to remove intrinsic noise and use inputs to update their estimates of environmental signals. 
Theorem \ref{theorem:langevin_replay} shows that, in the absence of inputs (e.g., sleep), these RNNs will perform Langevin sampling from the distribution of awake activity---thus, replaying awake activity.

\begin{theorem}
\label{theorem:optimal_update}
Given Assumption \ref{assumption:greedy}, the optimal update 
$\Delta \br^*(t) = \argmin_{\drt{2}} \widehat{\mathcal{L}}(t+\Delta t) + \argmin_{\drt{1}} \widehat{\mathcal{L}}(t+\Delta t) + \sigmaeta$
follows $\bs'(t)$ and the score function of $p(\brt)$:
\begin{equation}
    \Delta \br^*(t) =\bD^\dagger \bs'(t) \dt + \sigma_r^2 \dt \score + \sigmaeta
    \label{eq:delta_r_1_and_2_star}
\end{equation}
\end{theorem}
\begin{theorem}
\label{theorem:langevin_replay}
In the absence of input (quiescence), a noisy RNN already trained to path-integrate $\bst$ from $\bs'(t)$ will perform gradient ascent along the score function of $\brt$.
If $p(\brt)$ and $p(\bst)$ are stationary, then this ascent is Langevin sampling (Equation \ref{eq:overdamped_langevin}).
If the variance of $\sigmaeta$ is scaled by a factor of 2, then $p(\brt)$ is guaranteed to converge to the steady-state distribution $p(\br) = p(\bD^\dagger \bs)$.
\begin{equation}
\label{eq:rnn_langevin}
\text{If } \drt{} = \sigma_r^2 \dt \score + \sqrt{2} \sigmaeta, 
\enspace
\text{then} \lim_{t \rightarrow \infty} p(\br(t)) = p(\bD^\dagger \bs)
\end{equation}
\end{theorem}

\subsection{Replay Sampling vs. Other Types of Sampling}
\label{sec:replay_vs_other_sampling}

We study path-integrating noisy RNNs that, during quiescent states without external input, replay episodes of awake activity. 
Our RNNs generate trajectories in a causal, real-time fashion, one timestep at a time, unlike modern AI models which can generate all timesteps in a time-series simultaneously.

Crucially, our definition of sampling as time-series generation one timestep at a time should not be confused with the kind of Langevin sampling used by other works in computational neuroscience to explain how the brain may perform probabilistic calculations like Bayesian inference \citep{adaptation_accelerating, hamiltonian_brain, fast_sampling-based, natural_gradient_bayesian}.
We study time-series distributions: replay $p(\{ \brt \}_{t=0}^T)$, awake activity $p( \{ \bst \}_{t=0}^T )$ ($T$ is the final timestep), and how they compare---each time-series is a single sample from a time-series distribution with rigid temporal structure and non-stationarities ($p(\br(t_1)) \neq p(\br(t_2)) \, \forall \, t_1 \neq t_2$).
Meanwhile, inference networks use Langevin sampling as a means of generating sequences of activity $\{ \brt \}_{t=0}^T$ wherein each timestep is a separate sample from a static, stationary distribution: $\brt \sim p(\bs) \, \forall \, t$.
In inference networks, time simply becomes a means of gathering samples: with enough time, there are enough samples to compute static expectations, covariances, or posteriors via Monte Carlo.
Replay networks generate episodes of awake activity, while inference networks compute static statistics.
In \Cref{fig:setup_overview} we further illustrate the crucial distinction between these two notions of sampling.

Nonetheless, one might think existing techniques to accelerate Monte Carlo sampling in inference networks \citep{adaptation_accelerating, hamiltonian_brain, fast_sampling-based} should easily transfer to replay sampling networks, but adaptation clearly illustrates this is not the case.
\citet{adaptation_accelerating} show that adaptation accelerates Monte Carlo sampling over a static, constant input---they have an external input anchoring the network's target state, allowing negative feedback (adaptation) to rapidly destabilize the initial state without repelling the trajectory once it reaches the target.
In contrast, our networks internally generate replay without inputs.
Because no external drive exists to encourage reaching targets, negative feedback counteracts score-driven convergence towards learned attractors, causing the slowing we observe later. Adaptation accelerates input-clamped inference, but not necessarily internally-driven replay generation.

\subsection{Spike-Frequency Adaptation and Second-Order Dynamics}
\label{sec:bias_methods}

\paragraph{Spike-Frequency adaptation.}
Biological neurons can mitigate prolonged or low-frequency activity via negative feedback, i.e., spike-frequency \textit{adaptation} \citep{neural_adaptation, spike_frequency_adaptation}.
This feedback is important for describing \textit{in vivo} hippocampal activity and replay \citep{cell_assembly_sequences, nrem_sleep}, and in computational models of replay it encourages long replay trajectories (exploration) by preventing neural activity from getting stuck in attractor basins \citep{noisy_adaptation, dynamics_of_adaptive_canns, sequential_predictive_learning}.
Like \citet{sequential_predictive_learning}, we implement spike-frequency adaptation (normally defined for spiking neurons) in rate-based neural networks as negative feedback $\bct$ added to RNN activity $\brt$ (Equation \ref{eq:rnn_definition}) after training\footnote{Subtraction of the moving average $\bct$ also arises naturally from greedy minimization of $\loss + \frac{1}{2} \norm{\bct}^2$.}:  \begin{equation}
\label{eq:adaptation_definition}
\br(t + \dt) = f(\brt, \but, \sigmaeta) - \bct,
\quad
\Delta \bct = \frac{1}{\tau_a} (-\bct + b_a \brt)
\end{equation}

\paragraph{Biological basis of increasing adaptation during quiescence (sleep).}
In our main experiments, we train RNNs to path-integrate inputs without adaptation ($b_a = 0$) (an active, awake, state), and then during inference simulate the trained RNNs without inputs and with adaptation ($b_a \geq 0$) (a quiescent, sleep state).
Our justification for modulating adaptation strength comes from biology: the neuromodulators norepinephrine, acetylcholine (ACh), and to some extent serotonin, inhibit adaptation \citep{neurotransmitter_actions, selective_attenuation, serotonin_norepinephrine, norepinephrine_neuromodulator} and are less present during sleep than during wakefulness, a difference which is relevant for memory \citep{low_acetylcholine_pnas2004, low_acetylcholine_sleep2020, why_we_forget, neuropharmacology_of_sleep, serotonin_sleep_wake, effects_of_sleep}.
In other words, the brain already uses those neuromodulators to increase and decrease adaptation strength ($b_a$) during sleep and wakefulness, respectively.
\section{Second-Order Langevin Sampling for Neural Replay}

Based on the previous section, here we assume that trained RNNs have well-estimated the score function, and follow it during quiescent (internally-driven) activity to generate replay via Langevin dynamics.
Here we ask how modulating the dynamics of such RNNs affects or biases the replay distributions.
We see that adaptation (negative feedback) incurs a non-ideal form of second-order Langevin sampling, so we propose a complementary alternative that uses momentum and could be implemented in neurons via input differentiation.

\subsection{Adaptation as Underdamped Langevin Dynamics}
First we examine how introducing adaptation (negative feedback, as defined in \Cref{sec:bias_methods}) into the dynamics of a trained path-integrating RNN impacts its quiescent (internally-driven) dynamics. 
Our analysis gives an analytical account for why adaptation induces the exploration and slowness we observe later in replay from trained RNNs.
\Cref{prop:rnn_langevin_adaptation_1st_order} defines the first-order dynamics of a trained RNN with adaptation introduced after training, and \Cref{theorem:rnn_langevin_adaptation_2nd_order} re-writes them as a second-order dynamical system (derivation in \Cref{appendix:adaptation_sde}) to develop the insight in \Cref{remark:adaptation_conflict}.
\begin{proposition}
\label{prop:rnn_langevin_adaptation_1st_order}
Adding adaptation (\Cref{eq:adaptation_definition}) to a trained path-integrating RNN during quiescence (\Cref{eq:rnn_langevin}\footnote{
We study non-stationary systems, so we drop the $\sqrt{2}$ white noise scaling factor needed for stationary convergence.
}) incurs Langevin sampling with negative feedback:
\begin{equation}
\label{eq:rnn_langevin_adaptation_1st_order}
\drt{} = \sigmadt \, \score + \sigmaeta - \bct,
\quad
\Delta \bct = \frac{1}{\tau_a} (-\bct + b_a \brt)
\end{equation}
\end{proposition}
\begin{theorem}
\label{theorem:rnn_langevin_adaptation_2nd_order}
Adding adaptation to a trained path-integrating RNN during quiescence (\Cref{prop:rnn_langevin_adaptation_1st_order}) produces the following second-order stochastic dynamics (proof in \Cref{appendix:adaptation_sde}):
\begin{equation}
\label{eq:rnn_langevin_adaptation_2nd_order}
\begin{split}
\br''(t) = \left( 
\sigmadt \frac{d^2}{d \brt^2} \log p(\brt) - \frac{1}{\tau_a} \bm{I} 
\right) \br'(t)
- \frac{1}{\tau_a} \sigmaeta + \sigma_r \bm{\eta}'(t)
\\
+ \frac{1}{\tau_a} \Big( \underbrace{\sigmadt \score - b_a \brt}_{\textnormal{score conflicts with negative feedback}}
\Big)
\end{split}
\end{equation}
\end{theorem}
\begin{remark}
\label{remark:adaptation_conflict}
Interpreting \Cref{eq:rnn_langevin_adaptation_2nd_order} as underdamped (momentum) Langevin dynamics (\Cref{eq:underdamped_langevin_2nd_order}) reveals the effective score gradient $\sigmadt \score$ driving dynamics is shifted by $-b_a \brt$.
This negative feedback penalizes persistent activity, repelling activity from high-probability attractor states in $p(\brt)$.
For example, a Gaussian $\brt \sim \mathcal{N}(\bm{\mu}(t), \bm{\Sigma}(t))$\footnote{
For details on what $\bm{\mu}(t), \bm{\Sigma}(t)$ should be for a noisy RNN trained to path-integrate a signal $\bst$ distributed as an arbitrary Gaussian process, see \Cref{appendix:score_of_r(t)_s(t)}.
} 
has score $\score = \bm{\Sigma}(t)^{-1} (\bm{\mu}(t) - \brt)$, subtracting $b_a \brt$ from which pushes the Langevin dynamics away from their target $\bm{\mu}(t)$, decelerating convergence towards the target.
If $\bm{\mu}(t) = \bm{0}$, the score and $-\brt$ might align, accelerating convergence (depending on $\bm{\Sigma}(t)^{-1}$).
\end{remark}

Our \Cref{remark:adaptation_conflict} predicts that adaptation should in general push replay activity away from attractor states.
This should prevent activity from getting stuck in attractor basins, thus encouraging exploration during replay, but also decreasing the speed at which replay reaches new attractor basins.
We later observe empirically that adaptation induces these two phenomena---exploration and slowness---in replay, also matching our reasoning in \Cref{sec:replay_vs_other_sampling} that adaptation in spontaneous activity should generally slow, not accelerate, replay.
This raises the question of how one might design a biologically plausible network to accelerate replay via underdamped Langevin dynamics.

\subsection{Explicitly Underdamped Langevin Dynamics via Momentum}

To explicitly perform underdamped Langevin sampling in replay, we propose an alternative method based on Equation \ref{eq:underdamped_langevin_1st_order}.
It uses momentum\footnote{
Other RNNs have tried momentum \citep{momentumrnn}, but not for underdamped Langevin sampling or computational neuroscience.
}: a velocity term $\bvt$ accumulates previous values of $\brt$ when a friction term $\friction \in [0, 1]$ is below 1\footnote{
For a comparison of $\friction$ and the $\gamma$ term from Equations \ref{eq:underdamped_langevin_1st_order} and \ref{eq:underdamped_langevin_2nd_order}, see \Cref{appendix:deviations_langevin}.
}.
Here we define it and explain its biologically plausible implementation via dendritic input differentiation.
\begin{definition}
\label{definition:momentum}
We explicitly underdamp replay dynamics via \textit{momentum} governed by friction $\friction$.
When $\friction = 1$, dynamics revert to their default (overdamped, \Cref{eq:rnn_definition}):
\begin{equation}
\label{eq:underdamped_rnn}
\drt{} = \bm{v}(t + \dt) = (1 - \friction) \bvt + \underbrace{f(\brt, \but, \sigmaeta) - \brt}_{
\drt{} \text{ from overdamped case } (\friction = 1)
}
\end{equation}
\end{definition}

Although the dynamics above effectively implement momentum, it is unclear how a neural circuit might implement them.
By carefully examining the relationship between $\bm{v}(t + \dt)$ and $\br(t + \dt) = \brt + \bm{v}(t + \dt)$ in \Cref{appendix:plausible_momentum_proof}, we prove that a neural circuit could implement momentum by augmenting incoming (pre-synaptic) activity with the recent derivative thereof.
\begin{theorem}
\label{theorem:plausible_momentum}
The momentum dynamics from \Cref{definition:momentum} can be approximately implemented in a biologically plausible manner via input differentiation (proof in \Cref{appendix:plausible_momentum_proof}):
\begin{equation}
\label{eq:underdamped_plaus_rnn}
\br(t + \dt) = f(\underbrace{\brt + (1 - \friction) (\br(t - \dt) - \br(t - 2\dt))}_{\textnormal{effective incoming $\brt$}}, \but, \sigmaeta)
\end{equation}
\end{theorem}
\paragraph{Biological plausibility of input differentiation.}
We qualify the plausibility of \Cref{theorem:plausible_momentum} by answering three critical questions:
\begin{enumerate}
\item \textit{How can a single neuron differentiate its inputs?}
One straightforward way for a single neuron to calculate the recent derivative of its inputs is through dendritic delays: two different dendrites carry identical copies of input signals at different time delays, which are subtracted where the two branches converge.
Another way to differentiate inputs is through high-pass filtering.

\item \textit{How can the brain modulate the gain (strength) of the input differentiation?}
Neuromodulators in the brain like acetylcholine (ACh) can modulate the properties of neurons.
We propose they might also be able to modulate the strength of specific dendritic branches (i.e., ones computing inputs' recent derivatives), particularly if those branches have special receptors triggered by the neuromodulators.
For example, SK (small conductance calcium-activated potassium channels) channels and muscarinic M1 receptors, which ACh modulates \citep{m1_muscarinic_receptors, facilitation_long-term_potentiation}, could be concentrated in the derivative-computing dendritic branches.

\item \textit{How might the gain modulation correspond with active/quiescent (i.e., wake/sleep) cycles?}
As discussed in \Cref{sec:bias_methods}, the brain already changes levels of neuromodulators like ACh during wake/sleep cycles; this is how we justify increasing the adaptation strength $b_a$ during quiescent activity compared to awake activity.
\end{enumerate}
Overall, our aim is not to say concretely that the brain must implement momentum and must do so via input differentiation, but rather to suggest it as a topic of future experimental study.
Indeed, little is known about how the brain performs temporal compression of spontaneous activity.
\section{Numerical Results}

Now we examine how adaptation and biologically plausible momentum (Equations \ref{eq:adaptation_definition} and \ref{eq:underdamped_plaus_rnn}) bias replay distributions in simulated RNNs trained to path-integrate.
We first see that adaptation usually slows replay, while momentum quickens it.
Then we confirm that adaptation induces exploration, and crucially show that momentum does not prevent exploration, but rather complements it by increasing the distance that replay paths travel. 

\paragraph{Experiments.}
We have five tasks on which we train RNNs and then examine how introducing sampling mechanisms post-training (adaptation strength $b_a$ and friction $\friction$) affects replay (quiescent activity) statistics.
To encourage coherent replay, we train our RNNs to path-integrate with $\bm{s}'(t)$ periodically hidden (masked training,  \Cref{appendix:masked_training}).
We obtain quiescent activity by initializing the RNN hidden states (see \Cref{appendix:methods_expts}), then collecting activity over many unrolled timesteps, without external inputs but with internal noise still present.
Here we use ReLU RNNs, as justified in \Cref{appendix:hyperparameters}, but in Sections \ref{appendix:tanh} and \ref{appendix:non-plaus} we confirm that our results mostly hold in replay from tanh
RNNs\footnote{In tanh RNNs, we see momentum-induced compression, but for some reason not adaptation-induced slowing.}
and RNNs with non-plausible momentum.
In \Cref{appendix:methods} we define the replay statistics we measure and our RNN implementation details.
Additional measurements of replay statistics are in \Cref{appendix:more_results}.
Our five tasks are:
\begin{enumerate}
\item 1D Ornstein-Uhlenbeck (OU) process (\Cref{fig:1d_mean_std}): the optimal (with respect to \Cref{eq:loss_upper}) path-integrating RNN has a closed form (\Cref{eq:score_ou_main}), which we use in lieu of a trained RNN.
\item 2D T-maze and triangle (\Cref{fig:paths}): we simulate each direction of travel (T-maze: two; triangle: six) as a 2D OU process, and train a ReLU RNN to integrate 2D paths from every direction.
\item Rat trajectories (\Cref{fig:paths}): like \citet{sufficient_conditions}, we simulate 512 place cells that encode 2D directed (biased) or undirected (unbiased) random walks from RatInABox \citep{george2022ratinabox}, train a ReLU RNN to path-integrate place cell activity, and decode activity in 2D.
We use \textit{in silico} rat activity to mimic biological data because replay events \textit{in vivo} are sparse and difficult to extract from background noisy activity, therefore outside the scope of this work.
\end{enumerate}

\input{tikzfigs/paths_with_diagram}

\newcommand{\plausRtMinTmaze}{-28}
\newcommand{\plausRtMaxTmaze}{24}

\newcommand{\plausRtMinTriangle}{-30}
\newcommand{\plausRtMaxTriangle}{12}

\newcommand{\plausRtMinBiased}{-25}
\newcommand{\plausRtMaxBiased}{290}

\pgfplotsset{
    colormap={plausBwrTmaze}{
        color(\plausRtMinTmaze)=(blue)
        color(0)=(white)
        color(\plausRtMaxTmaze)=(red)},
    colormap={plausBwrTriangle}{
        color(\plausRtMinTriangle)=(blue)
        color(0)=(white)
        color(\plausRtMaxTriangle)=(red)},
    colormap={plausBwrBiased}{
        color(\plausRtMinBiased)=(blue)
        color(0)=(white)
        color(\plausRtMaxBiased)=(red)}
}

\begin{figure}[h]
\centering
\begin{tikzpicture}

\small

\begin{groupplot}[
    group style = {
        group size = 3 by 2,
        horizontal sep = 0.1\linewidth,
        vertical sep = 0.06\linewidth,
    },
    title style={yshift=-1.7ex,},
    width=0.24\linewidth,
    height=0.2\linewidth,
    xtick={0,1,2},
    xticklabels={$0$,$1$,$2$},
    xlabel style={yshift=0.5ex,},
    ytick={0,1,2,3},
    yticklabels={$1$,$0.9$,$0.8$,$0.7$},
    y dir=reverse,
    enlargelimits={abs=0.5},
    xmin=0, xmax=2,
    ymin=0, ymax=3,      
    ]

    \nextgroupplot[
        title={T-maze, median},
        ylabel=Friction $\friction$,
        point meta min=\plausRtMinTmaze,
        point meta max=\plausRtMaxTmaze,    
        colormap name=plausBwrTmaze,       
    ]
        \addplot [matrix plot*,point meta=explicit] file {csv/rt/plaus_med_rt_heatmap_tmaze.txt};

    \nextgroupplot[
        title={Triangle, median},
        point meta min=\plausRtMinTriangle,
        point meta max=\plausRtMaxTriangle,
        colormap name=plausBwrTriangle,
    ]
        \addplot [matrix plot*,point meta=explicit] file {csv/rt/plaus_med_rt_heatmap_triangle.txt};

    \nextgroupplot[
        title={Biased rat, median},
        point meta min=\plausRtMinBiased,
        point meta max=\plausRtMaxBiased,
        colormap name=plausBwrBiased,
        xmax=3,
        xtick={0,1,2,3},
        xticklabels={$0$, $1$, $2$, $3$},
        yticklabels={$1$, $0.7$, $0.5$, $0.3$}
    ]
        \addplot [matrix plot*,point meta=explicit] file {csv/rt/plaus_med_rt_heatmap_biased.txt};

    \nextgroupplot[
        title={T-maze, mean},
        point meta min=\plausRtMinTmaze,
        point meta max=\plausRtMaxTmaze,
        colormap name=plausBwrTmaze,  
        ylabel=Friction $\friction$,
        xlabel=Adaptation $b_a \times 100$,
        colorbar sampled, 
        colorbar horizontal,
        colorbar style={
            xtick={\plausRtMinTmaze, 0, \plausRtMaxTmaze},
            xticklabels={\textcolor{blue}{$\plausRtMinTmaze$}, $0$, \textcolor{red}{$+\plausRtMaxTmaze$}},            
            xticklabel style={font=\footnotesize}
        }        
    ]
        \addplot [matrix plot*,point meta=explicit] file {csv/rt/plaus_avg_rt_heatmap_tmaze.txt};

    \nextgroupplot[
        title={Triangle, mean},
        point meta min=\plausRtMinTriangle,
        point meta max=\plausRtMaxTriangle,    
        colormap name=plausBwrTriangle,    
        xlabel=Adaptation $b_a \times 100$,
        colorbar sampled, 
        colorbar horizontal,
        colorbar style={
            xtick={\plausRtMinTriangle, 0, \plausRtMaxTriangle},
            xticklabels={\textcolor{blue}{$\plausRtMinTriangle$}, $0$, \textcolor{red}{$+\plausRtMaxTriangle$}},
            xticklabel style={font=\footnotesize}
        }           
    ]
        \addplot [matrix plot*,point meta=explicit] file {csv/rt/plaus_avg_rt_heatmap_triangle.txt};

    \nextgroupplot[
        title={Biased rat, mean},
        point meta min=\plausRtMinBiased,
        point meta max=\plausRtMaxBiased,
        colormap name=plausBwrBiased,
        xmax=3,
        xtick={0,1,2,3},
        xticklabels={$0$, $1$, $2$, $3$},
        yticklabels={$1$, $0.7$, $0.5$, $0.3$},
        xlabel=Adaptation $b_a \times 100$,
        colorbar sampled, 
        colorbar horizontal,
        colorbar style={
            xtick={\plausRtMinBiased, 0, \plausRtMaxBiased},
            xticklabels={\textcolor{blue}{$\plausRtMinBiased \quad \quad$}, $0$, \textcolor{red}{$+\plausRtMaxBiased$}},
            xticklabel style={font=\footnotesize}
        }        
    ]
        \addplot [matrix plot*,point meta=explicit] file {csv/rt/plaus_avg_rt_heatmap_biased.txt};

\end{groupplot}

\node[inner sep=0] at (-1.7,-3.5) {\column{Change (\%) from\\awake statistics:}};

\node[inner sep=0] at (4.2, 2) {Replay reach time $(\downarrow)$ summary statistics};

\end{tikzpicture}
\caption{
\textbf{Momentum temporally compresses replay.}
We calculate how long it takes replay trajectories to reach their endpoints, relative to awake trajectories (\textcolor{blue}{blue} = faster, \textcolor{red}{red} = slower).
Biologically plausible momentum via input differentiation ($\friction < 1$) not only shortens this reach time, but makes it smaller than that of awake paths, \textit{temporally compressing} awake activity by up to $30\%$. 
We do not include unbiased rat trajectories because they do not have defined endpoints, but we do confirm in \Cref{appendix:more_results} that momentum quickens them.
The same trends hold in RNNs with tanh nonlinearities (\Cref{fig:tanh_reachtime_heatmaps_all}) and non-plausible momentum (\Cref{fig:non-plaus_reachtime_heatmaps_all}).
}
\label{fig:plaus_reachtime_heatmaps_all}
\end{figure}
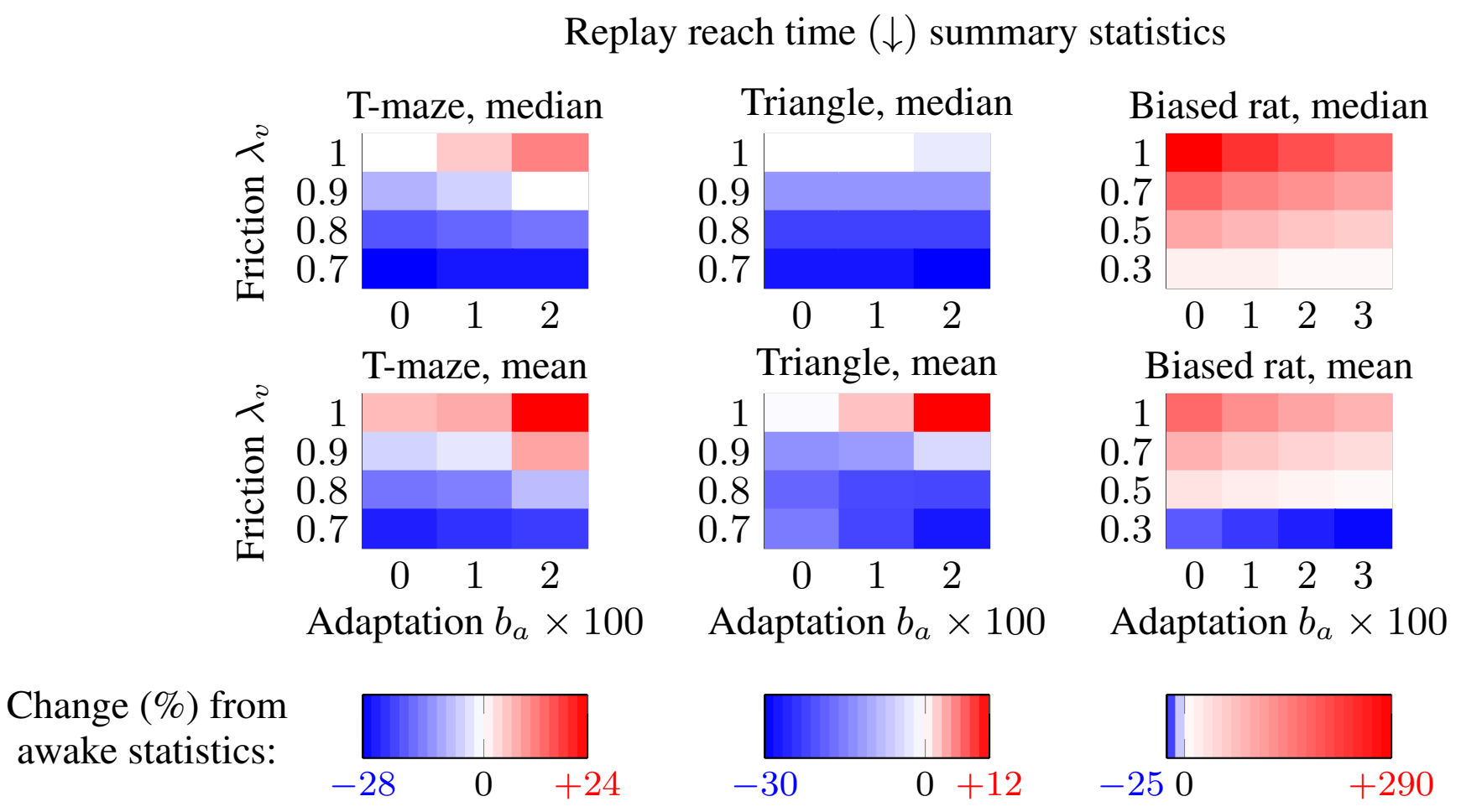

\paragraph{Adaptation slows replay, momentum accelerates it.}
We initially observe in Figures \ref{fig:1d_mean_std} and \ref{fig:paths} that adaptation slows replay by virtue of destabilizing attractor basins that replay trajectories target, while momentum via input differentiation accelerates the rate at which replay trajectories reach these targets. 
In \Cref{fig:plaus_reachtime_heatmaps_all} we quantify this observation by measuring how long it takes replay trajectories (in all but the unbiased rat task, since it lacks well-defined endpoint targets) to get within a certain radius of their ideal endpoints (see \Cref{appendix:methods} for details).
In the T-maze and triangle tasks, replay convergence to targets slows with adaptation, but quickens with momentum.
The notable exception is the biased rat task, wherein adaptation also accelerates convergence: this is because the biased path distributions all aim towards a common center, resembling Ornstein-Uhlenbeck processes with steady state mean $\mu = 0$---as explained in \Cref{remark:adaptation_conflict}, negative feedback can accelerate convergence towards a mean of $0$.
Nonetheless, in all three tasks, momentum can make replay trajectories faster in reaching targets than awake trajectories, performing \textit{temporal compression} of the latter by up to $30\%$.

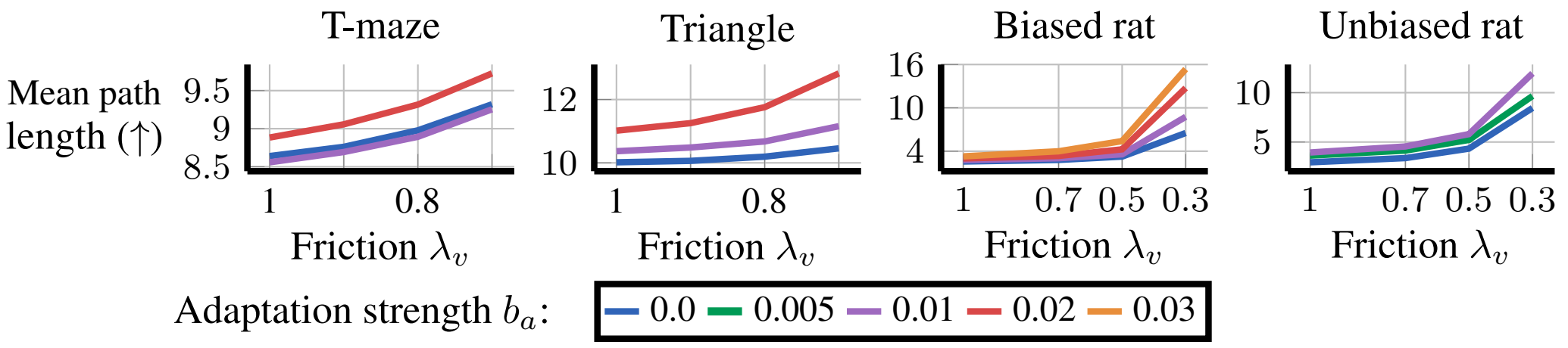
\begin{figure}[h!]
\centering
\begin{tikzpicture}

\pgfplotsset{/pgfplots/group/every plot/.append style = {
    ultra thick,
    legend style={
        legend columns=2,
        anchor=south west}
}};

\begin{groupplot}[
    group style = {
        group size = 4 by 1,
        horizontal sep = 0.05\linewidth,
        vertical sep = 0.05\linewidth},
    width = 0.28\linewidth,
    height = 0.18\linewidth,
    title style={yshift=-1.3mm,},
    x dir=reverse,   
    axis x line*=bottom,
    axis y line*=left,
    xlabel={Friction $\friction$},
    xlabel shift={-1mm},
    xticklabel style={font=\small},
    ylabel style={rotate=-90},
    ylabel shift={-3mm},
    yticklabel style={font=\small},
    grid,
    no marks,
    table/col sep=comma,
    cycle list name=mygradient
    ]

    \nextgroupplot[
        title=T-maze,
        ylabel=\column{\small Mean path\\length ($\uparrow$)},
        xtick={1, 0.9, 0.8, 0.7},
        xticklabels={1, , 0.8, },
        ]
    \foreach \bval in {0.0,0.01,0.02}{
        \addplot table[x=lambda_v, y={b_a=\bval}] {csv/exploration/plaus_exploration_tmaze_distances_avg.csv}; 
    }

    \nextgroupplot[
        title=Triangle,
        title style={yshift=-1mm,},
        xtick={1, 0.9, 0.8, 0.7},
        xticklabels={1, , 0.8, },]
    \foreach \bval in {0.0,0.01,0.02}{
        \addplot table[x=lambda_v, y={b_a=\bval}] {csv/exploration/plaus_exploration_triangle_distances_avg.csv};  
    }

    \nextgroupplot[
        title=Biased rat,
        xtick={1, 0.7, 0.5, 0.3},
        ytick={4, 10, 16}, 
        ymax=16,
        legend style={
            at={(-1.3,-1.6)},
            legend columns=5,
            row sep=-3pt,
            inner sep=1pt,
            font=\small}]
    \foreach \bval in {0.0,0.01,0.02,0.03}{
        \addplot table[x=lambda_v, y={b_a=\bval}] {csv/exploration/plaus_exploration_distance_biased.csv};  
        \addlegendentryexpanded{ \bval};
        \ifnum \fpeval{\bval == 0.0} = 1
            \addlegendimage{color=ForestGreen, line width=2pt};
            \addlegendentry{0.005};
        \fi
    }    

    \nextgroupplot[
        title=Unbiased rat,
        xtick={1,0.7,0.5,0.3},
        ] 
    \addplot[color=C0] table[x=lambda_v, y={b_a=0.0}] {csv/exploration/plaus_exploration_distance_unbiased.csv}; 
    \addplot[color=ForestGreen] table[x=lambda_v, y={b_a=0.005}] {csv/exploration/plaus_exploration_distance_unbiased.csv}; 
    \addplot[color=C1] table[x=lambda_v, y={b_a=0.01}] {csv/exploration/plaus_exploration_distance_unbiased.csv};

\end{groupplot} 

\node[inner sep=0] at (1,-1.25) {\column{Adaptation strength $b_a$:}};

\end{tikzpicture}
\caption{
\textbf{Momentum and adaptation both increase exploration via replay path length.}
Here we measure the cumulative path length of replay paths in replay trajectories across our four trained RNN tasks, so as to quantify how momentum and adaptation affect exploration in replay.
In all tasks we see the two mechanisms both increase replay path lengths.
The same trends hold in RNNs with tanh nonlinearities (\Cref{fig:tanh_exploration_distance}) and non-plausible momentum (\Cref{fig:non-plaus_exploration_distance}).
}
\label{fig:plaus_exploration_metrics}
\end{figure}

\input{tikzfigs/exploration_panels}

\paragraph{Momentum complements exploration from adaptation.}
We established that adaptation and momentum seem to counter each other with regards to speed.
However, replay \textit{in vivo} has both exploration and temporal compression.
Adaptation is known to induce exploration by destabilizing attractors \citep{sequential_predictive_learning}, but whether momentum will also counter that aspect of adaptation is unclear.
Upon further inspection of \Cref{fig:paths}, we see that the two mechanisms seem to complement each other in encouraging exploration.
To quantify this observed synergy shared by the two mechanisms, we devise several metrics for exploration (see \Cref{appendix:methods} for details), of which the two most salient ones are \textit{path lengths} and \textit{regions visited} of replay trajectories.
The former is the cumulative length of a replay path (not to be confused with the mean displacement of a replay trajectory from its origin, which we also measure in \Cref{fig:mean_displacement}), while the latter is a more semantic, task-specific measure of exploration in the T-maze and triangle tasks.
In Figures \ref{fig:plaus_exploration_metrics} and \ref{fig:plaus_region_count_avg} we confirm that both adaptation and momentum increase these metrics of exploration, with further confirmation in \Cref{appendix:more_results} via metrics including variance (\Cref{fig:time_variances}) and mean displacement (\Cref{fig:mean_displacement}).
While our several exploration metrics generally agree, we note that in the T-maze and triangle task (but not in the other two tasks), adaptation decreases mean displacement (\Cref{fig:mean_displacement}).
\section{Conclusions and Future Work}

By re-examining replay in path-integrating noisy RNNs as Langevin sampling, we established that first-order dynamics are insufficient to account for the speed and flexibility of biological replay.
We demonstrated analytically and empirically that spike-frequency adaptation acts as a form of non-ideal second-order sampling: while it successfully destabilizes local attractors to encourage exploration, it simultaneously introduces a dynamical drag that slows replay generation.
To resolve this exploration-speed trade-off, we introduced a biologically plausible momentum mechanism via dendritic input differentiation.
When combined, momentum counteracts slowing from adaptation, temporally compressing replay sequences while preserving, and even enhancing, multi-attractor spatial exploration.

Our findings suggest that temporal compression in networks like the hippocampus requires second-order circuit mechanisms.
While components of sequential predictive networks have been mapped to specific hippocampal subregions \citep{predictive_sequence_learning}, future work could refine our proposed mechanisms by incorporating biological constraints like Dale’s principle via explicit interneuron populations, by extending our Langevin framework to spike-based predictive learning models \citep{predictive_learning_rules, sequence_anticipation, learning_predictive_cognitive}, or by confirming our findings in more complex environments \citep{sequential_predictive_learning, honeycomb_maze}.

\subsection{Limitations \& Testable Predictions}
\label{sec:limitations}

Our theoretical framework carries several necessary abstractions. 
First, we assume the intrinsic noise in biological recurrent networks is Gaussian isotropic white noise to permit exact score estimation. 
Biological noise likely exhibits complex spatiotemporal correlations that could alter sampling dynamics. 
Second, directly comparing simulated replay to in vivo recordings remains challenging, as biological replay is sparse, sporadically embedded in background noise, and displays varying degrees of temporal compression.
Comparing our temporal compression results to observations \textit{in vivo} is also made challenging by a lack of consensus in the order of magnitude of temporal compression in the latter, which depending on modeling assumptions can range from 1--2x \citep{hippocampal_replay_experience} to 8--20x \citep{the_time_to_remember, consolidation_of_human, lee2002memory}.

Nonetheless, our framework yields explicit testable predictions for experimental neuroscience. Because neuromodulators like acetylcholine (ACh) modulate both spike-frequency adaptation and specific dendritic receptors (e.g., muscarinic M1 and SK channels), we propose that the drop in these neuromodulators during sleep not only disinhibits adaptation but may also facilitate dendritic input differentiation.
We predict that targeted pharmacological increase of these specific neuromodulators during quiescence would severely attenuate the temporal compression of replay events.
We also predict that second-order modifiers (momentum, adaptation) will improve models of replay as diffusive dynamics \citep{hippocampal_reactivation_random, drift-diffusion_dynamics}.


\section{Acknowledgements}
This work was supported by NSF grants CCF-1911094 and IIS-1730574; ONR grants N00014-23-1-2714, N00014-24-1-2225, and MURI N00014-20-1-2787; AFOSR grant FA9550-22-1-0060; DOE grant DE-SC0020345; DOI grant 140D0423C0076; and a Vannevar Bush Faculty Fellowship, ONR grant N00014-18-1-2047.


\paragraph{Code and data availability.}
All code to reproduce our results is in 
\url{https://github.com/Josuelmet/rnn-replay-second-order/tree/main}


\bibliographystyle{unsrtnat}
\bibliography{_bib,_bib_nanda}


\newpage
\appendix
\section*{Supplementary Material}
\section{Methods}
\label{appendix:methods}

\subsection{Hyperparameters Shared Across Experiments}
\label{appendix:hyperparameters}

\subsubsection{Masked Training}
\label{appendix:masked_training}
Denoisers and autoencoders benefit from \textit{masked training}, wherein some regions of input data are set to zero before model processing \citep{survey_on_masked}. 
\citet{sequential_predictive_learning} introduce masked training for path-integration by periodically masking the input $\but$ (the observation of $\bs'(t)$): only at every $k$-th timestep does the RNN observe a nonzero input (Equation \ref{eq:masked_training}).
\begin{equation}
    \label{eq:masked_training}
    \but = \begin{cases}
        \bs'(t), & t \text{ mod } k = 0 \\
        \bm{0}, & \mathrm{otherwise}
    \end{cases}
\end{equation}
\citet{sequential_predictive_learning} found that masked training makes replay sequences more coherent and makes manifolds of neural activity more similar to the spatial layout of the environment.
We found that masked training improves replay stability, so we use it (with $k \geq 3$) in training all RNNs.

\subsubsection{Activation Function}

In our work, we have mostly used RNNs with ReLU or leaky ReLU activation functions (except in \Cref{appendix:tanh}); our reasons for doing so are threefold.
The first is that ReLU or ReLU-like activations, unlike tanh activations, do not always saturate. This can mitigate vanishing gradients in RNNs (although care must be taken to avoid exploding gradients), but more importantly we have observed that this reduces the strength of fixed attractors learned by path-integrating RNNs, which is useful for exploration (in \Cref{appendix:tanh} we observe that replay in saturating RNNs explores less in the triangle task).
The second is that, compared to RNNs with multiplicative gating interactions (e.g., GRUs and LSTMs), ReLU RNNs are more biologically plausible insofar as they can be easily interpreted as the nonnegative, sparse firing rates of spiking networks.
The third is that most other previous works in sequential predictive coding tend to use ReLU activations, and have shown no significant differences in results when using more complex nonlinearities:
\begin{itemize}
\item \citet{predictive_coding_consequence}, some of the first authors to report the emergence of predictive representations emerging in RNNs, use ReLU activations.
\item \citet{sufficient_conditions} primarily use ReLU networks, and report similar results with GRU networks.
\item \citet{sequential_predictive_learning} use ReLU activations in conjunction with layer normalization, but report similar results when layer normalization is removed. 
\item \citet{predictive_sequence_learning} use tanh for most results, but when they care about the emergence of sparse, localized, nonnegative activations (i.e., place cells), they use ReLU activations.
\item \citet{sorscher2019unified, conformal_isometry}, and \citet{excitatory_inhibitory_recurrent} were all interested in the emergence of grid cells in predictive representations and primarily used ReLU or ReLU-like activations; those that tried using other nonlinearities did not observe notable differences in results.
\item \citet{learning_grid_cells}, who were also interested in grid cells, use both ReLU and tanh activations, and also report no notable differences between the two.
\item \citet{sequential_memory} use linear networks, and notice no significant difference when using tanh activations.
\end{itemize}

\subsection{Experiments}
\label{appendix:methods_expts}

All experiments were conducted in PyTorch using a single consumer GPU (NVIDIA Titan X Pascal 12 GB).

\subsubsection{Ornstein-Uhlenbeck}

Awake trajectories $s(t)$ are simulated with $\dt = 0.02, \sigma_s = 0.1, \sigma_0 = 0.2, \theta = 2, \mu = 5$ for $T = 100$ iterations.
We used \Cref{eq:score_ou_main} as the deterministic component of $\drt{}$ since a 1D Ornstein-Uhlenbeck process admits a closed-form expression for $\frac{d}{dr(t)} \log p(r(t))$.
In \Cref{fig:1d_mean_std}, momentum (underdampening) corresponds to $\friction=0.5$, while adaptation corresponds to $b_a = 0.01, \tau_a = 20$.

\subsubsection{T-maze and Triangle}

\paragraph{Task description.}
In the 2D T-maze and triangle tasks, we simulate directed random walks (Ornstein-Uhlenbeck processes) along several directions, and train the RNN to path-integrate these walks from their velocities.
In the T-maze task, there are two directions of travel: both start from the origin and go up, then one goes left and the other goes right (orange and purple, respectively, in \Cref{fig:paths}).
Meanwhile, in the triangle task there are six directions: denoting the three corners of the equilateral triangle as $A = (0,0), B = (1,0)$, and $C = (1/2, \sqrt{3}/2)$, respectively, these directions are $\overrightarrow{AB}, \overrightarrow{BC}, \overrightarrow{CA}, \overrightarrow{AC}, \overrightarrow{CB}, \overrightarrow{BA}$, shown in \Cref{fig:paths} as blue, red, gray, pink, green, and orange, respectively.
Awake and replay paths last $100$ timesteps, unless exploration is being measured, in which replay paths are allowed to last $400$ timesteps.

\paragraph{Architecture and training.}
For both tasks, we used shallow leaky-ReLU RNNs with linear output projections, as described in \Cref{appendix:discretization}.
In the T-maze task, we used 20 hidden neurons, while in the triangle task, we used 40.
We use masking difficulty $k=3$ in a progressive curriculum.
Triangle-task RNNs are trained with $k=1$ for $20,000$ epochs, then at $k=2$ and $k=3$ for $5,000$ epochs each; T-maze-task RNNs are trained likewise, but with fewer epochs ($12,000$ at $k=1$ and $5,000$ at $k=2$ and $k=3$ each).
We train 9 RNN seeds for the triangle task, and 10 for the T-maze task.

\paragraph{Hidden state initialization.}
In each task, multiple directions start from the same point in space (for example, $\overrightarrow{AB}$ and $\overrightarrow{AC}$ in the triangle task), so we add onto initial hidden states some random vectors, orthogonal to the 2D output projection, specific to each direction.
For example, hidden states for $\overrightarrow{AB}$ and $\overrightarrow{AC}$ paths are both initialized to start near $A$, but $\overrightarrow{AB}$ paths also start with a fixed random vector $\bm{\eta}_{AB}$ added to their initial hidden states, whereas $\overrightarrow{AC}$ have a different fixed random vector $\bm{\eta}_{AC}$ added to their initial hidden states.
This notion of initializing hidden states in directions orthogonal to output projections has been previously discussed in computational neuroscientific contexts \citep{preparatory_activity}, and recent evidence suggests that memories (which replay is similar to) have uniquely identifiable, output-orthogonal patterns of neural activation \citep{barcoding_episodic}.

\subsubsection{Rat Place Cell Trajectories}

Our unbiased (undirected random walks within a 2D box) and biased (directed random walks that head towards the center of a 2D box) rat place cell trajectory experiments are identical to those of \citep{sufficient_conditions}, with 10 different ReLU RNN seeds per task, but with two modifications:
\begin{enumerate}
\item We add masked training, with difficulties $k=3$ in the biased task and $k=6$ in the unbiased task.
We use a higher masking difficulty in the unbiased task to encourage longer replay trajectories; at lower values of $k$, unbiased replay paths are much shorter than awake trajectories.
In Langevin sampling terms, we conjecture this is because the score function $\score$ is fairly weak since unbiased trajectories are uniformly distributed\footnote
{
This relative weakness of the unbiased activity score function (especially compared to the biased activity score function, where trajectories all go towards the center of a 2D box) is also why we use relatively small adaptation strengths $b_a$ in simulating replay from unbiased-task RNNs.
}.
\item We use a slower, but more detailed, decoder. \citep{sufficient_conditions} decode position at a given timestep as the average position of the top 3 most active place cells, which is fast but effectively quantizes decoded positions.
We use this method to initialize our positions, but then we optimize our positions via gradient descent to minimize the mean-squared error between observed place cell activity and the place cell activity that would correspond to these optimized positions (this is made possible by our knowledge of the exact place cell activity function). 
Our procedure takes much longer, but results in truly continuous replay trajectories, avoiding any significant quantization.
\end{enumerate}

\subsection{Metrics}

\subsubsection{Speed: Reach Times}

In the T-maze, triangle, and biased rat tasks, awake trajectories have clearly defined endpoints that they reach.
Replay trajectories also aim for these endpoints, although second-order dynamics modifiers like adaptation and momentum (underdampening) affect how quickly or how closely replay paths reach them.
We quantify these changes by measuring the timesteps required for replay paths to get within a certain radius of their endpoints (for example, the timesteps required for $\overrightarrow{AB}$ paths to get within a 0.1$|\overrightarrow{AB}|$ radius of $B$).

\subsubsection{Exploration: Path Lengths}

One way we quantify exploration in replay paths is through path length.
We calculate this simply as the sum of velocity magnitudes.

\subsubsection{Exploration: Regions Visited}

Another way we quantify exploration in the T-maze and triangle tasks is through \textit{regions visited}.
As illustrated in \Cref{fig:plaus_region_count_avg}, \textit{regions} are portions of input space where all points are closest to the same endpoint.
In the triangle task, there are 3 endpoints (which all also act as initial points), while in the T-maze there are 2 endpoints and 1 shared starting point, which for the purposes of region assignment we also consider an ``endpoint''.
Thus, in each task there are 3 endpoints, and thus 3 regions.
Awake trajectories go from one region to another and stay there, but replay trajectories might proceed to visit more regions.
We define a ``visit'' as a contiguous presence within a single region for at least 10 timesteps; naturally, exploratory replay sequences can re-visit previously visited regions, which is how their ``regions visited'' metric can exceed 3.

\subsubsection{Exploration: Additional Metrics}
\begin{figure}
\centering
\includegraphics[width=0.9\linewidth]{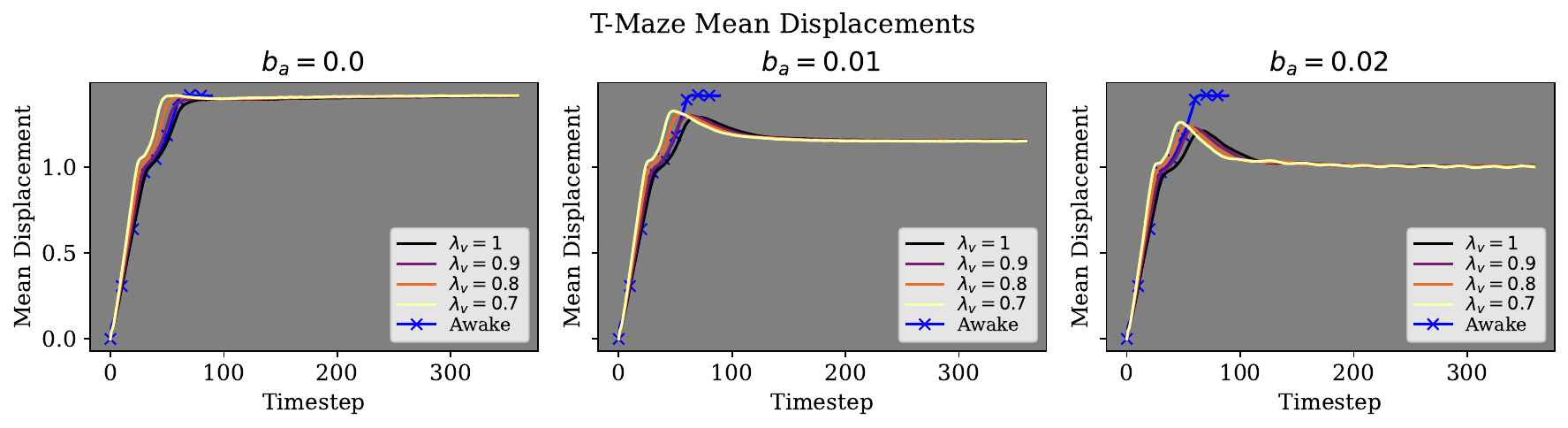}
\includegraphics[width=0.9\linewidth]{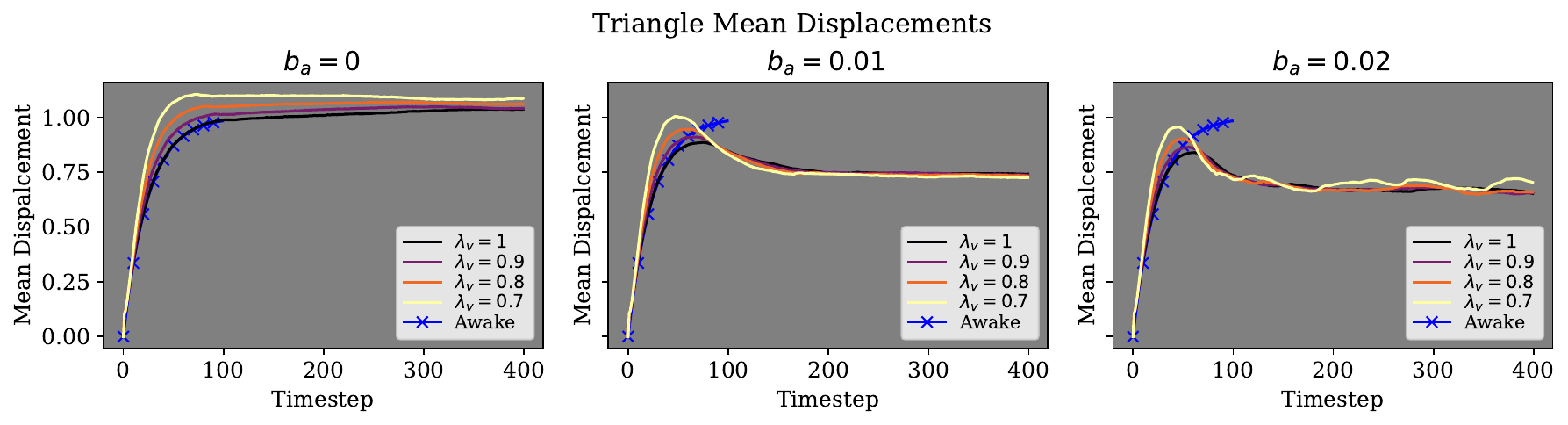}
\includegraphics[width=0.9\linewidth]{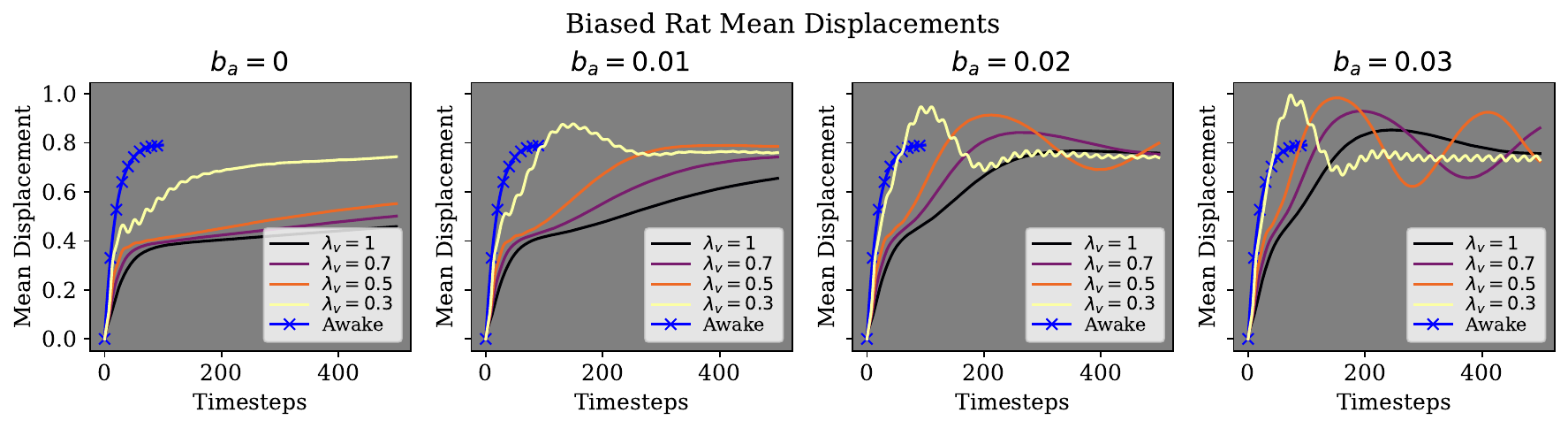}
\includegraphics[width=0.9\linewidth]{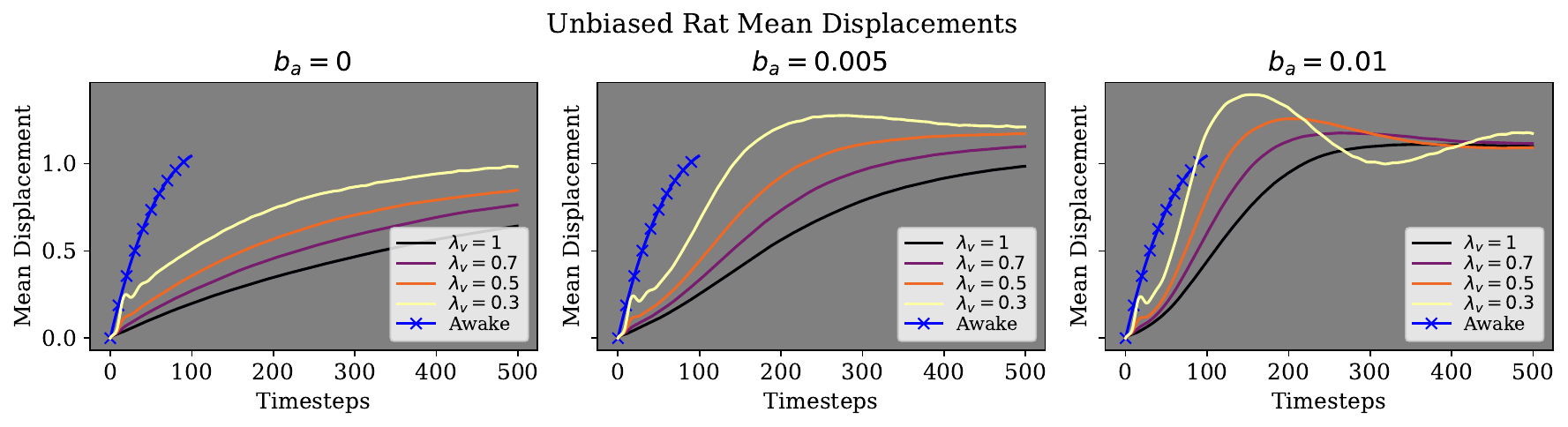}
\caption{
\textbf{Momentum ($\friction < 1$) increases mean displacement of replay trajectories, especially at early timesteps.}
Similarly to \citet{flexible_modulation}, here we analyze the \textit{mean displacement}, or distance, of replay trajectories from their starting points as a function of time.
For reference, mean displacement over time is also plotted for awake trajectories.
In exploration experiments, we simulate replay for $4\times$ the duration of awake trajectories.
Higher mean displacements over time generally correlate with increased path length and exploration.
Note that awake activity is always calculated with $b_a=0, \friction=1$, hence why awake statistics are the same across plots within the same task.
}
\label{fig:mean_displacement}
\end{figure}

\begin{figure}
\centering
\includegraphics[width=0.9\linewidth]{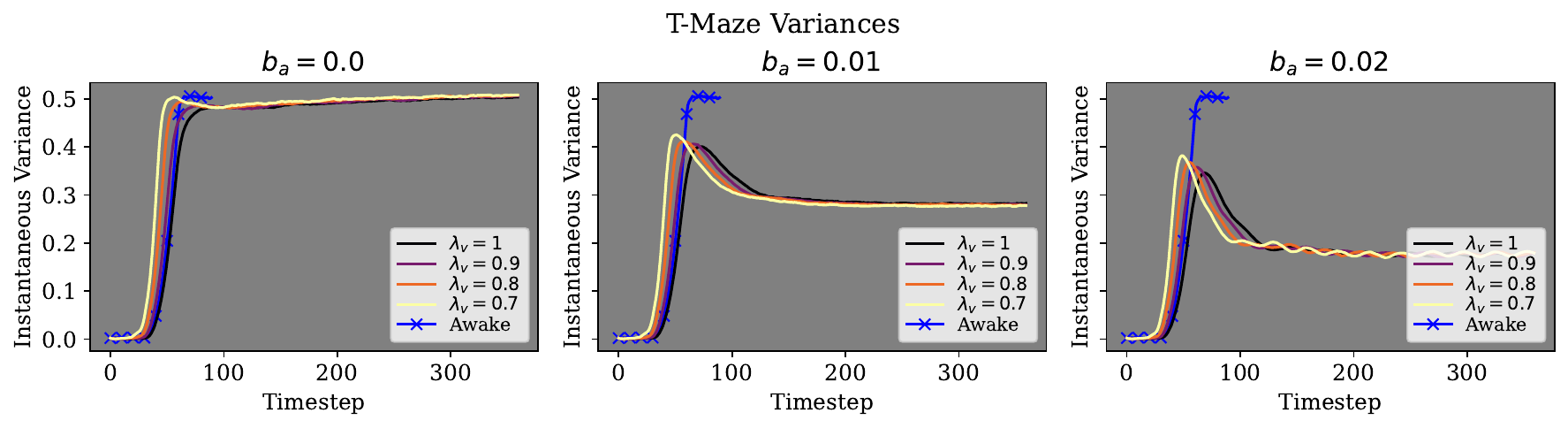}
\includegraphics[width=0.9\linewidth]{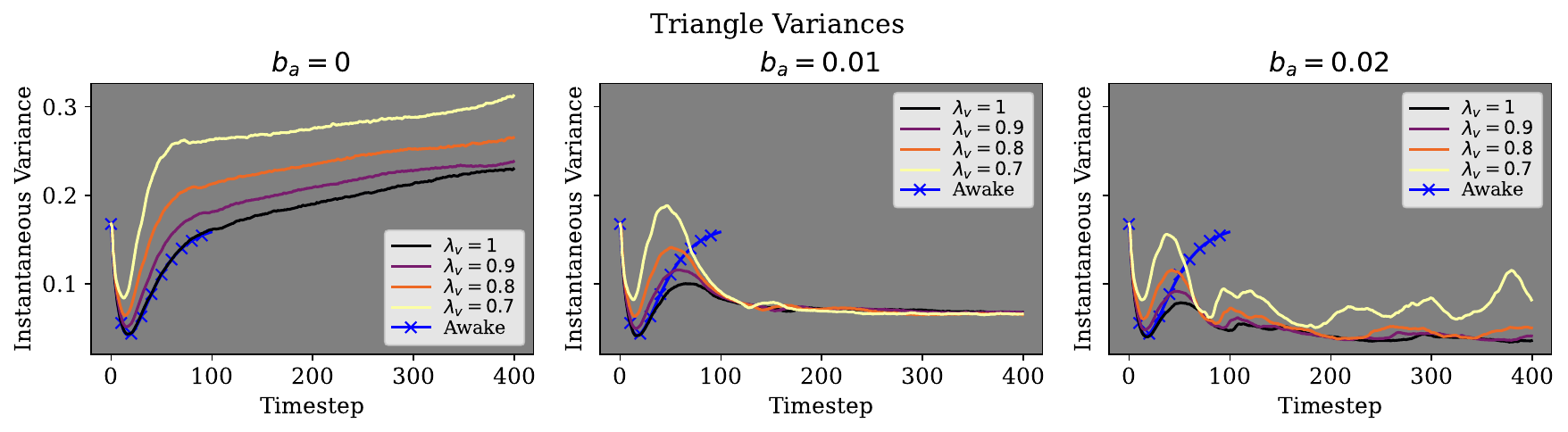}
\includegraphics[width=0.9\linewidth]{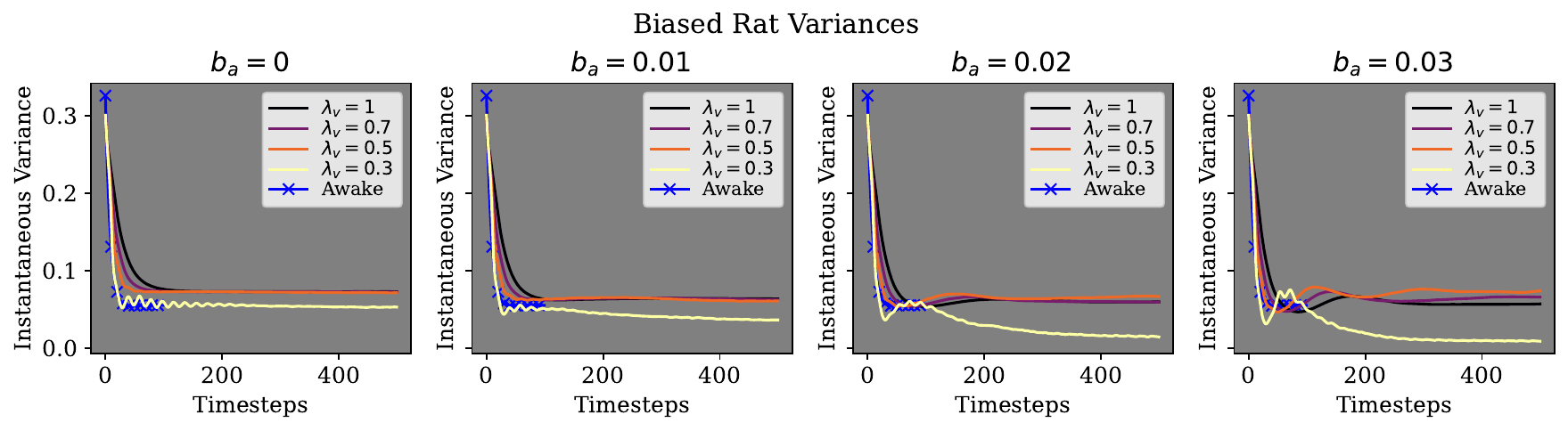}
\includegraphics[width=0.9\linewidth]{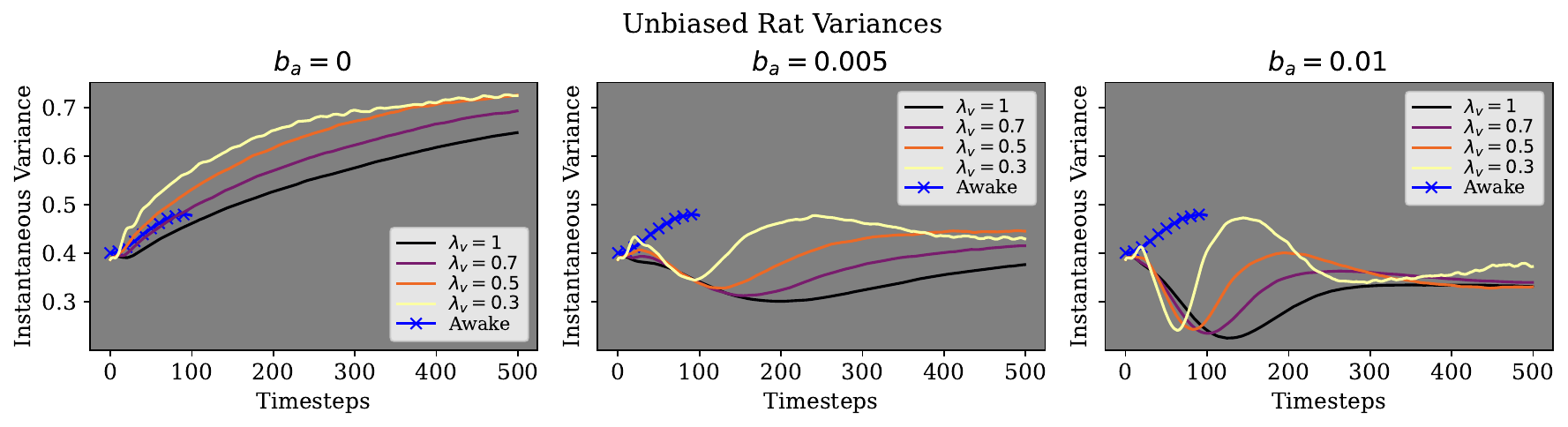}
\caption{
\textbf{Momentum ($\friction < 1$) increases replay trajectory variances.}
Like in \Cref{fig:mean_displacement}, here we analyze the variance of trajectories at each timestep.
Trajectories are 2D, so the variances plotted above are simply the averages of the variances along each coordinate.
Momentum increasing variance is another confirmation that it complements adaptation-induced exploration.
}
\label{fig:time_variances}
\end{figure}

For completeness, we add two other measures of exploration: \emph{mean displacement}, proposed by \citet{flexible_modulation}, and variance.
They are simply defined as $\mathbb{E}_{p(\brt)} [ |\brt - \br(0)| ]$ and $\mathbb{E}_{p(\brt)} [\mathrm{Var}(\brt)]$.
Higher mean displacements and variances generally correlate with increased path length and exploration.
We plot them in \Cref{fig:mean_displacement} and \Cref{fig:time_variances}.

\subsubsection{Wasserstein Distance}
\label{appendix:wasserstein_distance}
We seek to compare distributions of trajectories. 
In most of our tasks, these are objects of dimension $T \times 2$, where $T$ is the number of timesteps (recall that T-maze and triangle tasks are already in 2D, while rat trajectories are analyzed after being projected into 2D space).
To compare such high-dimensional distributions, we use Wasserstein distances \citep{statistical_aspects_wasserstein}:
\begin{itemize}
\item In the T-maze and triangle tasks, we first calculate the Wasserstein distance between awake and replay paths belonging to the same direction (e.g., $\overrightarrow{AB}$ in the triangle task).
Since paths along the same direction should resemble, if not obey, Gaussian processes, these paths should approximately be normally distributed, and so we can use the closed-form equation for the Wasserstein distance between two normal distributions.
After computing Wasserstein distances for each direction, we take the average as our final distance.
KL divergence might have also worked in this task, but in practice the covariance matrices of the $(T \times 2)$-dimensional distributions were singular.
\item In the rat experiments, decoded 2D trajectories are not easily decomposed into groups of Gaussian processes, so instead we apply sliced Wasserstein distances \citep{sliced_radon_wasserstein} to compare the $(T \times 2)$-dimensional distributions.
Sliced Wasserstein distances are essentially calculated by taking many random projections of two distributions onto 1D, where Wasserstein distances have a closed form.
Computing KL divergences instead would have been challenging, since rat path distributions are high-dimensional and do not admit straightforward estimations of probability density.
\end{itemize}

\subsection{Discretization and Implementation of Noisy RNNs}
\label{appendix:discretization}

While biological neural networks are continuous-time systems, RNNs are in practice implemented discretely.
In order to incorporate second-order processes like adaptation and momentum in RNNs, we must discretize them, which we initially did as follows, and call ``non-plausible momentum'':
\begin{align}
\Delta \widetilde{\br}(t) &= f(\brt, \but, \sigmaeta) - \brt
\\
\bm{c}(t + \dt) &= \bct + \frac{1}{\tau_a} (-\bct + b_a \brt)
\\
\bm{v}(t + \dt) &= (1 - \friction) \bvt + \Delta \widetilde{\br}(t)
\\
\br(t + \dt) &= \brt -\bct + \bm{v}(t + \dt)
\end{align}
Note that, unlike $\bm{c}(t + \dt)$ and $\bm{v}(t + \dt)$, $\br(t + \dt)$ depends on another term calculated at $t + \dt$. This is a symplectic Euler discretization of the second-order dynamics, which we employ to ensure the stability of interactions between $\brt$ and its momentum $\bvt$.
The negative feedback $\bct$ is much more stable, so its discretization goes unchanged.

If the adaptation strength $b_a = 0$, then there is no adaptation ($\bct = \bm{0}$).
As for our \textit{friction} term $\friction \in [0, 1]$, if $\friction = 1$, then there is no momentum since $\bvt = \drt{} + \sigmaeta$ no longer accumulates previous values of $\bvt$.
In other words, the friction term $\friction$ allows for smooth interpolation between overdamped and momentum sampling.

While we initially worked with the dynamics above (see \Cref{appendix:non-plaus} for the results thereof), they use a biologically implausible momentum via $\bvt$.
Therefore, we re-define the dynamics above with a plausible version of momentum via input differentiation to get our final, main dynamics (derivation in  \Cref{appendix:plausible_momentum_proof}):
\begin{align}
\bm{c}(t + \dt) &= \bct + \frac{1}{\tau_a} (-\bct + b_a \brt)
\\
\br(t + \dt) &= f(\brt + (1 - \friction) (\br(t - \dt) - \br(t - 2\dt)), \but, \sigmaeta) - \bct
\end{align}

We train and sample from ReLU RNNs either with or without leaky integration: $f(\brt, \but, \sigmaeta) = \kappa \brt +  \mathrm{ReLU}(\bm{W}_r \brt + \bm{W}_{in} \but + \sigmaeta)$ (where $\kappa \in [0, 1]$ is either $0$ or learnable) unless otherwise stated; we justify our choice of ReLU activation function in \Cref{appendix:hyperparameters}, but do ensure to verify that our results generalize to other activations functions in \Cref{appendix:tanh}.
In the absence of directed inputs $\but$, $f(\brt, \but, \sigmaeta)$ should be roughly equivalent to the noisy score function from Equation \ref{eq:rnn_langevin}.

Across all trained RNN experiments we use $b_a = 0, \friction=1$ for awake activity (i.e., training).
When generating replay, we always use $\tau_a = 50$, and either $T=100, 400,$ or $500$ timesteps, depending on whether we are measuring speed and fidelity ($T=100$ in Figures \ref{fig:paths}, \ref{fig:plaus_reachtime_heatmaps_all}, \ref{fig:plaus_wd}) or exploration ($T=400$ for T-maze and triangle, $T=500$ for rat tasks in Figures \ref{fig:plaus_exploration_metrics}, \ref{fig:plaus_region_count_avg}).

\subsection{Deviations from Traditional Langevin Sampling}
\label{appendix:deviations_langevin}

We must note a few minor distinctions between our replay RNNs and traditional Langevin sampling.

\paragraph{Stationarity.}
We treat neural replay as a sequence of events, and are thus interested in the joint distribution of replay activity at all timesteps $p\left(\{\brt\}_{t=1}^T\right)$.
This distribution, however, is not necessarily stationary across $t$. Stationarity would imply that $p(\brt)$ has no need or intention of traversing along any meaningful path.
While a path-integrating RNN does perform gradient ascent along $\log p(\brt)$, it does so in a piecewise manner along $t$ rather than jointly along all $t$ simultaneously.
This variation on gradient ascent, in combination with the non-stationarity of $p(\brt)$, means that Langevin sampling guarantees do not hold.
For this reason, we make modifications to our RNNs that under stationary Langevin sampling theory might seem unprincipled.

\paragraph{Momentum (explicit underdampening).}
The friction term $\friction$ applied to the velocity $\bvt$ is subtly different from $\gamma$ in Equations \ref{eq:underdamped_langevin_1st_order} and \ref{eq:underdamped_langevin_2nd_order}.
$\friction \in [0, 1]$ can be interpreted as trying to map $\gamma \in [0, \infty)$ onto $[0, 1]$.
Like $\gamma \rightarrow \infty$, $\friction \rightarrow 1$ removes any dependency of $\bm{v}(t + \dt)$ on $\bvt$.

\paragraph{Noise scaling.}
RNN replay, unlike Langevin sampling, is sensitive to the variance of noise used.
We found that omitting the $\sqrt{2}$ factor in front of $\sigmaeta$ worked best.
For this reason, we also omitted $\friction$ as a scaling term on $\sigmaeta$.
If $\friction$ is allowed to modulate $\sqrt{2} \sigmaeta$, we found that it is easy to show improvements in replay fidelity when $\friction < 1$, but we found these to come from noise scaling rather than from momentum as a  mechanism.

\paragraph{Relation to diffusion models.}
Throughout this work we treat RNNs as generative models using a variant of Langevin sampling.
Expressive generative models, in particular \textit{diffusion models} whose activity resembles Langevin sampling, have received much attention from the machine learning community \citep{self_consuming, boomerang, diffusion_models_comprehensive, diffusion_models_in_vision, diffusion-lm, chatgpt_is_not, generative_modeling_by_estimating_gradients}.
The fundamental differences between our RNNs and diffusion models are twofold.
The first is that each timestep of a generated replay trajectory is generated sequentially and with only one effective step along the gradient of log-likelihood, whereas a diffusion model would generate all timesteps of a path simultaneously and with many steps along the gradient of log-likelihood.
The second is that diffusion models use time-varying noise levels (annealed dynamics) and are explicitly conditioned on time as an input, whereas our RNNs use the same noise level across time and are never explicitly conditioned on time.
\section{Second-Order Stochastic Differential Equation Derivations}

\subsection{Adaptation as a Second-Order Stochastic Differential Equation}
\label{appendix:adaptation_sde}

We start by abstracting \Cref{eq:rnn_langevin_adaptation_1st_order}, the quiescent dynamics of a trained path-integrating RNN with adaptation introduced after training, with the following substitutions:
the score function $\sigmadt \score \rightarrow g(\brt)$, scaled noise $\sigmaeta \rightarrow \bm{\varepsilon}(t)$, and the adaptation parameters $\frac{1}{\tau_a} \rightarrow a$ and $\frac{b_a}{\tau_a} \rightarrow b$.
With these, we re-write \Cref{eq:rnn_langevin_adaptation_1st_order} into a single second-order stochastic differential equation as follows:
\begin{align}
\br'(t) &= g(\br(t)) - \bct + \bm{\varepsilon}(t), 
\quad 
\bm{c}'(t) = -a\bct + b\brt
\\
\br''(t) &= g'(\brt) \br'(t) - \bm{c}'(t) + \bm{\varepsilon}'(t)
\\
&= g'(\brt) \br'(t) - (-a\bct + b \brt) + \bm{\varepsilon}'(t), 
\quad
\bct = -\br'(t) + g(\brt) - \bm{\varepsilon}(t)
\\
&= g'(\brt) \br'(t) - (-a[-\br'(t) + g(\brt) - \bm{\varepsilon}(t)] + b \brt) + \bm{\varepsilon}'(t)
\\
&= g'(\brt) \br'(t) - a\br'(t) + ag(\brt) - a\bm{\varepsilon}(t) - b \brt + \bm{\varepsilon}'(t)
\\
&= (g'(\brt) - a) \br'(t) + (ag(\brt) - b\brt) - a\bm{\varepsilon}(t) + \bm{\varepsilon}'(t)
\end{align}
Replacing the substitution variables with their original values yields the following (\Cref{eq:rnn_langevin_adaptation_2nd_order}):
\begin{equation}
\begin{split}
\br''(t) = \left( 
\sigmadt \frac{d^2}{d \brt^2} \log p(\brt) - \frac{1}{\tau_a} \bm{I} 
\right) \br'(t)
\\
+ \frac{1}{\tau_a} \left(
\sigmadt \score - b_a \brt
\right) 
\\
- \frac{1}{\tau_a} \sigmaeta + \sigma_r \bm{\eta}'(t)
\end{split}
\end{equation}

\subsection{Biologically Plausible Momentum Derivation}
\label{appendix:plausible_momentum_proof}

Here we derive our biologically plausible momentum mechanism via input differentiation.
We start with the definitions of our (discretized) dynamics from \Cref{appendix:discretization}, assuming there is no adaptation ($b_a = 0$, so $\bct = \bm{0}$):
\begin{align}
\bm{v}(t + \dt) &= (1 - \friction) \bvt + f(\brt, \but, \sigmaeta) - \brt
\\
\br(t + \dt) &= \brt + \bm{v}(t + \dt)
\end{align}
Then, we substitute one equation into the other, canceling out the $\brt$ terms:
\begin{align}
\br(t + \dt) &= \brt + (1 - \friction) \bvt + f(\brt, \but, \sigmaeta) - \brt
\\
&= (1 - \friction) \bvt + f(\brt, \but, \sigmaeta)
\end{align}
By definition, $\bvt = \brt - \br(t - \dt)$.
Therefore:
\begin{equation}
\br(t + \dt) = (1 - \friction) (\brt - \br(t - \dt)) + f(\brt, \but, \sigmaeta)
\end{equation}

Momentum is therefore equal to adding the scalar-weighted recent velocity $(1 - \friction)(\brt - \br(t - \dt))$ to the calculation of the next timestep of activity $\br(t + \dt)$.
This is implausible for a neuron to do.
However, it is plausible for neurons to differentiate their inputs and use these derivatives as separate inputs, so we propose to move the differentiation from the \textit{outputs} of neurons (pre-synaptic) to their \textit{inputs} (post-synaptic), essentially applying momentum to the $\brt$ term of $\br(t + \dt)$ rather than $\br(t + \dt)$ itself:
\begin{align}
\br(t + \dt) &= f(\br_{in}(t), \but, \sigmaeta), \enspace \text{where}
\\
\br_{in}(t) &= \brt + (1 - \friction) (\br(t - \dt) - \br(t - 2\dt))
\end{align}
Momentum is therefore implementable in a biologically plausible manner via a specific variant of input differentiation.
Of course, if $\friction = 1$, $\br_{in}(t) = \br(t)$.
\section{Score Functions of Gaussian Distributions and Processes}
\label{appendix:gaussian_scores}

For any matrix calculus involved, we use denominator layout.

\subsection{Score of a Multivariate Gaussian Distribution}

Let's suppose $\br \sim \mathcal{N}(\bmu, \bSigma)$.
If $\br \in \mathbb{R}^d$, then:
\begin{align}
    p(\br) &= \frac{1}{\sqrt{(2 \pi)^d |\bSigma|}} \exp \left( -\frac{1}{2} (\br - \bmu)^T \bSigma^{-1} (\br - \bmu) \right)
    \\
    \log p(\br) &\propto -\frac{1}{2} (\br - \bmu)^T \bSigma^{-1} (\br - \bmu)
    \\
    \nabla_{\br} \log p(\br) &= -\frac{1}{2} ((\bSigma^{-1})^T + \bSigma^{-1}) (\br - \bmu)
    \\
    &= -\bSigma^{-1} (\br - \bmu)
    \label{eq:gaussian_score}
    \\
    &= -\sigma^{-2}(r - \mu) \enspace \mathrm{if} \enspace r \in \mathbb{R}
\end{align}

\subsection{Score of \texorpdfstring{$r(t)$}{r(t)} for Any Gaussian Process \texorpdfstring{$s(t)$}{s(t)}}
\label{appendix:score_of_r(t)_s(t)}

Recall that $p(\brt | \bst) \sim \mathcal{N}(\bDdag \bst, \, \bI \sigmadt)$ from Equation \ref{eq:p_r|s}.
If we suppose that $\bst$ is normally distributed with mean and covariance $\bmu_{\bst}, \bSigma_{\bst}$, then we can obtain $p(\brt)$:
\begin{equation}
    p(\brt) \sim \mathcal{N}(\bDdag \bmu_{\bst}, \, \bI \sigmadt + \bDdag \bSigma_{\bst} (\bDdag)^T),
\end{equation}
which we can plug into Equation \ref{eq:gaussian_score} to get $\score$:
\begin{equation}
    \score =
    -\left(
    \bI \sigmadt +
    \bDdag \bSigma_{\bst} (\bDdag)^T
    \right)^{-1}
    \left(
    \brt - \bDdag \bmu_{\bst}
    \right)
\end{equation}
Moreover, we can use the above score function to calculate the optimal $\Delta \brt$ in Equation \ref{eq:delta_r_1_and_2_star}:
\begin{equation}
\begin{split}
    \Delta \br^*(t) &=
    \sigmadt \left(
    \bI \sigmadt +
    \bDdag \bSigma_{\bst} (\bDdag)^T
    \right)^{-1}
    \left(
    -\brt + \bDdag \bmu_{\bst}
    \right)
    \\
    &+ \bDdag \bs'(t) \dt + \sigmaeta
\end{split}
\end{equation}
Some properties of the leakage matrix $\sigmadt \left( \bI \sigmadt + \bDdag \bSigma_{\bst} (\bDdag)^T \right)^{-1}$ include:
\begin{enumerate}
    \item The covariance matrix $\bSigma_{\bst}$ is positive semidefinite (PSD): all its eigenvalues are $\geq 0$.
    \item $\bDdag \bSigma_{\bst} (\bDdag)^T$ is also PSD
    \footnote{
    Proof: If $\bm{B}$ is PSD, then $x^T \bm{ABA}^T x = (\bm{A}^Tx)^T \bm{B} (\bm{A}^T x) = v^T \bm{B} v \geq 0$.
    }, symmetric, and therefore diagonalizable.
    \item The eigenvalues of $(\bI \sigmadt + \bDdag \bSigma_{\bst} (\bDdag)^T)^{-1}$ are thus all $\leq (\sigmadt)^{-1}$
    \footnote{
    Proof: For diagonalizable $\bm{A}$, the $i$-th eigenvalue of $(\lambda \bI + \bm{A})^{-1} = (\bm{Q}(\lambda \bI + \bm{\Lambda})\bm{Q}^{-1})^{-1}$ is equal to $(\lambda + \bm{\Lambda}_{ii})^{-1}$, which can be no larger than $\lambda^{-1}$ if $\bm{A}$ is PSD.
    }. 
    \item The eigenvalues of $ \sigmadt (\bI \sigmadt + \bDdag \bSigma_{\bst} (\bDdag)^T)^{-1}$ are thus all $\leq 1$.
    \item If the off-diagonal entries of the leakage matrix above are sufficiently small in magnitude, then all the diagonal entries should be less than 1, as justified by the Gershgorin Circle Theorem.
    In fact, if the leakage matrix is diagonal, then it must have all values less than 1 (which could be achieved via sigmoid functions or perhaps spectral normalization).
    \item As for interpretation, smaller leakage eigenvalues means higher eigenvalues of $\bDdag \bSigma_{\bst} (\bDdag)^T$, or essentially, more noise.
    The maximum determinant of the leakage matrix is 1, when there is essentially no noise in $\bs(t)$.
\end{enumerate}

\clearpage
\subsection{Score of a few Markov Gaussian Processes}
\label{appendix:score_markov_gaussian}

\paragraph{Wiener Processes.}
One simple stochastic process is the Wiener process, which in terms of navigation represents an undirected random walk ($\theta = 0$).
Assuming $s_w(0) = 0$, then $s_w(t) \sim \mathcal{N}(0, \sigma_s^2 t)$, and therefore $p(r_w(t)) \sim \mathcal{N}(0, \sigma_s^2 t + \sigmadt)$ from Equation \ref{eq:p_r|s}, producing the following score:
\begin{equation}
\sigmadt \, \nabla_{r_w(t)} \log p(r_w(t)) = \sigmadt \frac{-r_w(t)}{\sigma_s^2 t + \sigmadt} 
\end{equation}
Even from a simple Wiener process, we observe that the per-timestep optimal score is not constant with respect to $t$: at $t=0$, it equals $-r_{w}(t)$, while as $t$ approaches $\infty$, it approaches $0$.

\paragraph{Ornstein-Uhlenbeck Processes.}

Now we incorporate non-zero leakage ($\theta > 0$) to describe a directed random walk navigating from an arbitrary starting point $s_{ou}(0)$ towards a mean destination $\mu$.
If $p(s_{ou}(0)) \sim \mathcal{N}(0, \sigma_0^2)$, then $p(s_{ou}(t)) \sim \mathcal{N} \left(\mu (1 - e^{-\theta t}), \frac{\sigma_s^2}{2 \theta} (1 - e^{-2 \theta t}) + \sigma_0^2 e^{-2\theta t} \right)$, and the score function is:
\begin{equation} 
\label{eq:score_ou_main}
\sigmadt \, \nabla_{r_{ou}(t)} \log p(r_{ou}(t)) = \sigmadt \frac{
-(r_{ou}(t) - \mu(1 - e^{-\theta t}))
}{
\frac{\sigma_s^2}{2\theta}(1 - e^{-2\theta t}) + \sigma_0^2 e^{-2\theta t} + \sigmadt
}
\end{equation}
\section{Additional Results}
\label{appendix:more_results}

Here we present additional results or figures that supplement those of the main text.

\begin{figure}[h]
\centering
\begin{tikzpicture}

\pgfplotsset{/pgfplots/group/every plot/.append style = {
    ultra thick,
    legend style={
        legend columns=2,
        anchor=south west}
}};

\begin{groupplot}[
    group style = {
        group size = 4 by 1,
        horizontal sep = 0.05\linewidth,
        vertical sep = 0.05\linewidth},
    width = 0.28\linewidth,
    height = 0.18\linewidth,
    title style={yshift=-1.3mm,},
    x dir=reverse,   
    axis x line*=bottom,
    axis y line*=left,
    xlabel={Friction $\friction$},
    xlabel shift={-1mm},
    xticklabel style={font=\small},
    ylabel style={rotate=-90},
    ylabel shift={-3mm},
    yticklabel style={font=\small},
    grid,
    no marks,
    table/col sep=comma,
    cycle list name=mygradient
    ]

    \nextgroupplot[
        title=T-maze,
        ylabel=\column{\small Wasserstein\\distance ($\downarrow$)},
        ylabel style={
            rotate=90,
            yshift=2mm
        },
        xtick={1, 0.9, 0.8, 0.7},
        xticklabels={1, , 0.8, },
        ]
    \foreach \bval in {0.0,0.01,0.02}{
        \addplot table[x=lambda_v, y={b_a=\bval}] {csv/wd/plaus_wd_tmaze.csv}; 
    }

    \nextgroupplot[
        title=Triangle,
        title style={yshift=-1mm,},
        xtick={1, 0.9, 0.8, 0.7},
        xticklabels={1, , 0.8, },]
    \foreach \bval in {0.0,0.01,0.02}{
        \addplot table[x=lambda_v, y={b_a=\bval}] {csv/wd/plaus_wd_triangle.csv};  
    }

    \nextgroupplot[
        title=Biased rat,
        xtick={1, 0.7, 0.5, 0.3},
        ytick={0.05, 0.1}, 
        yticklabels={0.05, 0.1},
        legend style={
            at={(-1.3,-1.6)},
            legend columns=5,
            row sep=-3pt,
            inner sep=1pt,
            font=\small}]
    \foreach \bval in {0.0,0.01,0.02,0.03}{
        \addplot table[x=lambda_v, y={b_a=\bval}] {csv/wd/plaus_wds_biased.csv};  
        \addlegendentryexpanded{ \bval};
        \ifnum \fpeval{\bval == 0.0} = 1
            \addlegendimage{color=ForestGreen, line width=2pt};
            \addlegendentry{0.005};
        \fi
    }    

    \nextgroupplot[
        title=Unbiased rat,
        xtick={1,0.7,0.5,0.3},
        ] 
    \addplot[color=C0] table[x=lambda_v, y={b_a=0.0}] {csv/wd/plaus_wds_unbiased.csv}; 
    \addplot[color=ForestGreen] table[x=lambda_v, y={b_a=0.005}] {csv/wd/plaus_wds_unbiased.csv}; 
    \addplot[color=C1] table[x=lambda_v, y={b_a=0.01}] {csv/wd/plaus_wds_unbiased.csv};

\end{groupplot} 

\node[inner sep=0] at (1,-1.25) {\column{Adaptation strength $b_a$:}};

\end{tikzpicture}
\caption{
Wasserstein distance (\Cref{appendix:wasserstein_distance}) of replay trajectories compared to awake trajectories.
Versions for RNNs with tanh activations and non-plausible momentum are in Figures \ref{fig:tanh_wd} and \ref{fig:non-plaus_wd}.
}
\label{fig:plaus_wd}
\end{figure}

\section{Results with Non-Plausible Momentum}
\label{appendix:non-plaus}

Here we reproduce our main findings, but with non-plausible momentum (\Cref{appendix:discretization}) instead of our plausible momentum via input differentiation.

\newcommand{\nonPlausRtMinTmaze}{-28}
\newcommand{\nonPlausRtMaxTmaze}{24}
\newcommand{\nonPlausRtMinTriangle}{-30}
\newcommand{\nonPlausRtMaxTriangle}{12}
\newcommand{\nonPlausRtMinBiased}{-65}
\newcommand{\nonPlausRtMaxBiased}{290}

\pgfplotsset{
    colormap={nonPlausBwrTmaze}{
        color(\nonPlausRtMinTmaze)=(blue)
        color(0)=(white)
        color(\nonPlausRtMaxTmaze)=(red)},
    colormap={nonPlausBwrTriangle}{
        color(\nonPlausRtMinTriangle)=(blue)
        color(0)=(white)
        color(\nonPlausRtMaxTriangle)=(red)},
    colormap={nonPlausBwrBiased}{
        color(\nonPlausRtMinBiased)=(blue)
        color(0)=(white)
        color(\nonPlausRtMaxBiased)=(red)}
}

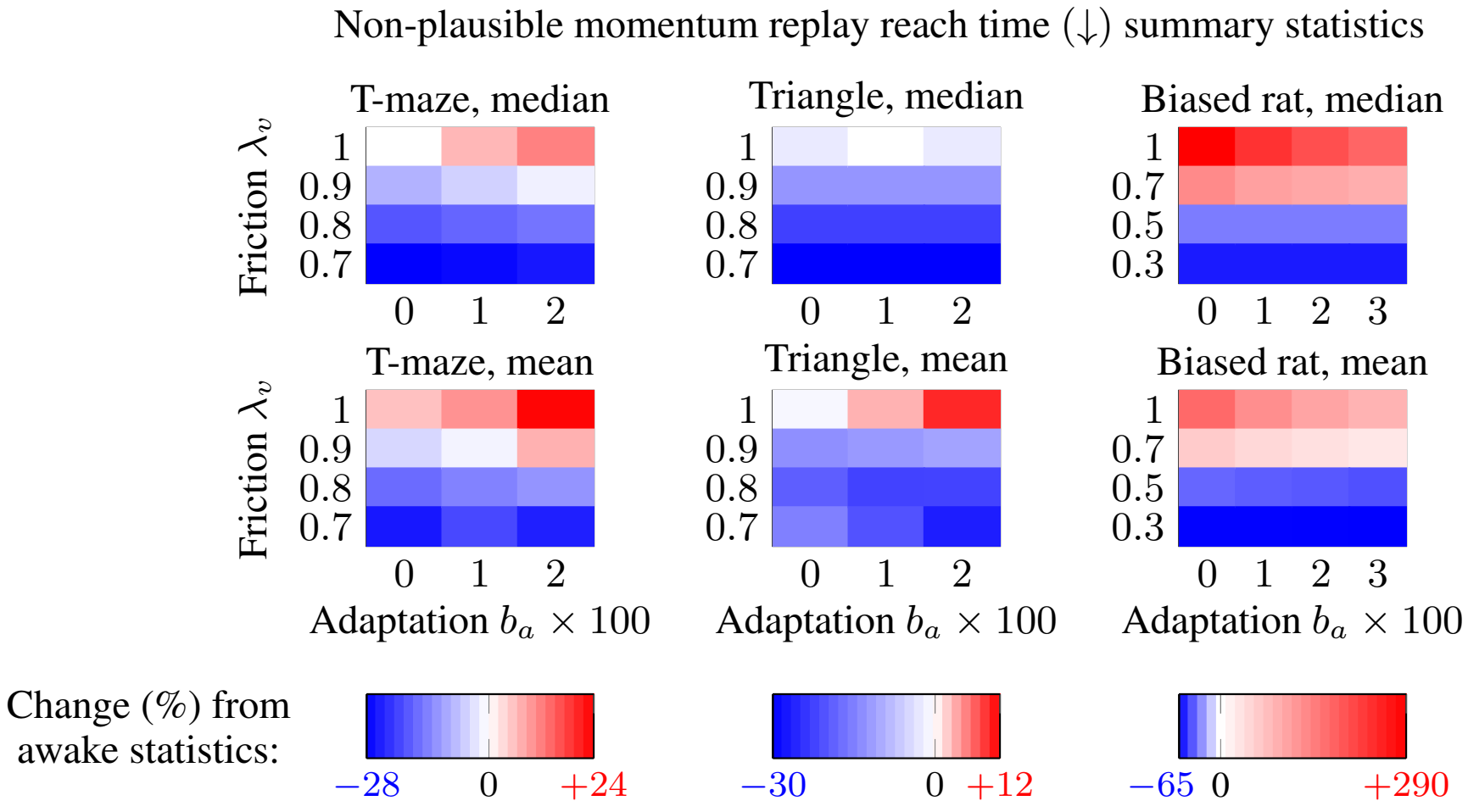
\begin{figure}[h]
\centering
\begin{tikzpicture}

\small

\begin{groupplot}[
    group style = {
        group size = 3 by 2,
        horizontal sep = 0.1\linewidth,
        vertical sep = 0.06\linewidth,
    },
    title style={yshift=-1.7ex,},
    width=0.24\linewidth,
    height=0.2\linewidth,
    xtick={0,1,2},
    xticklabels={$0$,$1$,$2$},
    xlabel style={yshift=0.5ex,},
    ytick={0,1,2,3},
    yticklabels={$1$,$0.9$,$0.8$,$0.7$},
    y dir=reverse,
    enlargelimits={abs=0.5},
    xmin=0, xmax=2,
    ymin=0, ymax=3,      
    ]

    \nextgroupplot[
        title={T-maze, median},
        ylabel=Friction $\friction$,
        point meta min=\nonPlausRtMinTmaze,
        point meta max=\nonPlausRtMaxTmaze,    
        colormap name=nonPlausBwrTmaze,       
    ]
        \addplot [matrix plot*,point meta=explicit] file {csv/rt/med_rt_heatmap_tmaze.txt};

    \nextgroupplot[
        title={Triangle, median},
        point meta min=\nonPlausRtMinTriangle,
        point meta max=\nonPlausRtMaxTriangle,
        colormap name=nonPlausBwrTriangle,
    ]
        \addplot [matrix plot*,point meta=explicit] file {csv/rt/med_rt_heatmap_triangle.txt};

    \nextgroupplot[
        title={Biased rat, median},
        point meta min=\nonPlausRtMinBiased,
        point meta max=\nonPlausRtMaxBiased,
        colormap name=nonPlausBwrBiased,
        xmax=3,
        xtick={0,1,2,3},
        xticklabels={$0$, $1$, $2$, $3$},
        yticklabels={$1$, $0.7$, $0.5$, $0.3$}
    ]
        \addplot [matrix plot*,point meta=explicit] file {csv/rt/med_rt_heatmap_biased.txt};

    \nextgroupplot[
        title={T-maze, mean},
        point meta min=\nonPlausRtMinTmaze,
        point meta max=\nonPlausRtMaxTmaze,
        colormap name=nonPlausBwrTmaze,  
        ylabel=Friction $\friction$,
        xlabel=Adaptation $b_a \times 100$,
        colorbar sampled, 
        colorbar horizontal,
        colorbar style={
            xtick={\nonPlausRtMinTmaze, 0,\nonPlausRtMaxTmaze},
            xticklabels={\textcolor{blue}{$\nonPlausRtMinTmaze$}, $0$, \textcolor{red}{$+\nonPlausRtMaxTmaze$}},
            xticklabel style={font=\footnotesize}
        }        
    ]
        \addplot [matrix plot*,point meta=explicit] file {csv/rt/avg_rt_heatmap_tmaze.txt};

    \nextgroupplot[
        title={Triangle, mean},
        point meta min=\nonPlausRtMinTriangle,
        point meta max=\nonPlausRtMaxTriangle,    
        colormap name=nonPlausBwrTriangle,    
        xlabel=Adaptation $b_a \times 100$,
        colorbar sampled, 
        colorbar horizontal,
        colorbar style={
            xtick={\nonPlausRtMinTriangle, 0, \nonPlausRtMaxTriangle},
            xticklabels={\textcolor{blue}{$\nonPlausRtMinTriangle$}, $0$, \textcolor{red}{$+\nonPlausRtMaxTriangle$}},
            xticklabel style={font=\footnotesize}
        }           
    ]
        \addplot [matrix plot*,point meta=explicit] file {csv/rt/avg_rt_heatmap_triangle.txt};

    \nextgroupplot[
        title={Biased rat, mean},
        point meta min=\nonPlausRtMinBiased,
        point meta max=\nonPlausRtMaxBiased,
        colormap name=nonPlausBwrBiased,
        xmax=3,
        xtick={0,1,2,3},
        xticklabels={$0$, $1$, $2$, $3$},
        yticklabels={$1$, $0.7$, $0.5$, $0.3$},
        xlabel=Adaptation $b_a \times 100$,
        colorbar sampled, 
        colorbar horizontal,
        colorbar style={
            xtick={\nonPlausRtMinBiased, 0, \nonPlausRtMaxBiased},
            xticklabels={\textcolor{blue}{$\nonPlausRtMinBiased \quad$}, 0, \textcolor{red}{$+\nonPlausRtMaxBiased$}},
            xticklabel style={font=\footnotesize}
        }        
    ]
        \addplot [matrix plot*,point meta=explicit] file {csv/rt/avg_rt_heatmap_biased.txt};

\end{groupplot}

\node[inner sep=0] at (-1.7,-3.5) {\column{Change (\%) from\\awake statistics:}};

\node[inner sep=0] at (4, 2) {Non-plausible momentum replay reach time $(\downarrow)$ summary statistics};

\end{tikzpicture}
\caption{
\textbf{Momentum temporally compresses replay in RNNs with non-plausible momentum.}
This is another version of \Cref{fig:plaus_reachtime_heatmaps_all}, but now with non-plausible momentum (see \Cref{appendix:discretization}).
See \Cref{fig:tanh_reachtime_heatmaps_all} for a version with plausible momentum in tanh RNNs.
}
\label{fig:non-plaus_reachtime_heatmaps_all}
\end{figure}

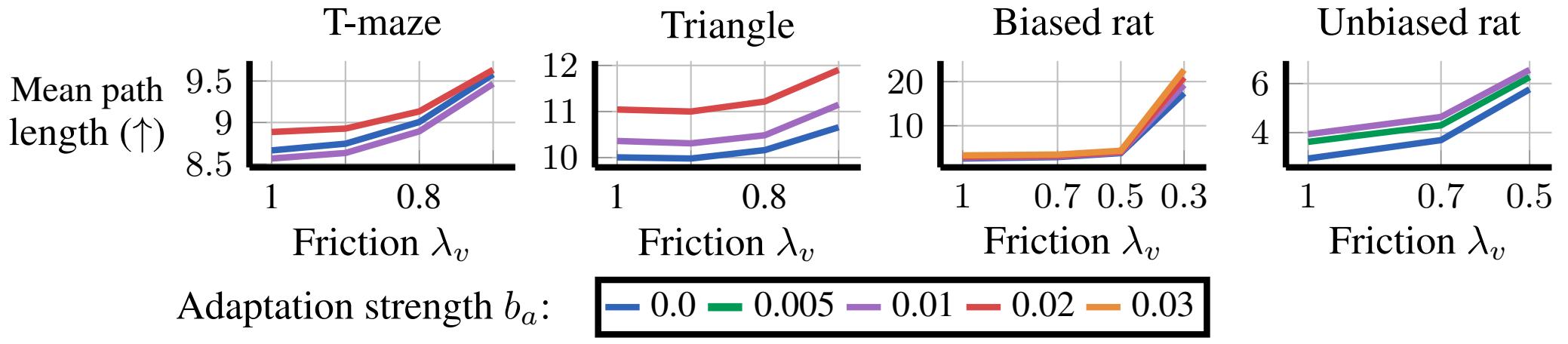
\begin{figure}[h!]
\centering
\begin{tikzpicture}

\pgfplotsset{/pgfplots/group/every plot/.append style = {
    ultra thick,
    legend style={
        legend columns=2,
        anchor=south west}
}};

\begin{groupplot}[
    group style = {
        group size = 4 by 1,
        horizontal sep = 0.05\linewidth,
        vertical sep = 0.05\linewidth},
    width = 0.28\linewidth,
    height = 0.18\linewidth,
    title style={yshift=-1.3mm,},
    x dir=reverse,   
    axis x line*=bottom,
    axis y line*=left,
    xlabel={Friction $\friction$},
    xlabel shift={-1mm},
    xticklabel style={font=\small},
    ylabel style={rotate=-90},
    ylabel shift={-3mm},
    yticklabel style={font=\small},
    grid,
    no marks,
    table/col sep=comma,
    cycle list name=mygradient
    ]

    \nextgroupplot[
        title=T-maze,
        ylabel=\column{\small Mean path\\length ($\uparrow$)},
        xtick={1, 0.9, 0.8, 0.7},
        xticklabels={1, , 0.8, },
        ]
    \foreach \bval in {0.0,0.01,0.02}{
        \addplot table[x=lambda_v, y={b_a=\bval}] {csv/exploration/exploration_tmaze_distances_avg.csv}; 
    }

    \nextgroupplot[
        title=Triangle,
        title style={yshift=-1mm,},
        xtick={1, 0.9, 0.8, 0.7},
        xticklabels={1, , 0.8, },]
    \foreach \bval in {0.0,0.01,0.02}{
        \addplot table[x=lambda_v, y={b_a=\bval}] {csv/exploration/exploration_triangle_distances_avg.csv};  
    }

    \nextgroupplot[
        title=Biased rat,
        xtick={1, 0.7, 0.5, 0.3},
        legend style={
            at={(-1.3,-1.6)},
            legend columns=5,
            row sep=-3pt,
            inner sep=1pt,
            font=\small}]
    \foreach \bval in {0.0,0.01,0.02,0.03}{
        \addplot table[x=lambda_v, y={b_a=\bval}] {csv/exploration/exploration_distance_biased.csv};
        \addlegendentryexpanded{ \bval};
        \ifnum \fpeval{\bval == 0.0} = 1
            \addlegendimage{color=ForestGreen, line width=2pt};
            \addlegendentry{0.005};
        \fi
    }    

    \nextgroupplot[
        title=Unbiased rat,
        xtick={1,0.7,0.5},
        ] 
    \addplot[color=C0] table[x=lambda_v, y={b_a=0.0}] {csv/exploration/exploration_distance_unbiased.csv}; 
    \addplot[color=ForestGreen] table[x=lambda_v, y={b_a=0.005}] {csv/exploration/exploration_distance_unbiased.csv}; 
    \addplot[color=C1] table[x=lambda_v, y={b_a=0.01}] {csv/exploration/exploration_distance_unbiased.csv};

\end{groupplot} 

\node[inner sep=0] at (1,-1.25) {\column{Adaptation strength $b_a$:}};

\end{tikzpicture}
\caption{
\textbf{Momentum and adaptation both increase exploration via replay path length in RNNs with non-plausible momentum.}
This is another version of \Cref{fig:plaus_exploration_metrics}, but now with non-plausible momentum (see \Cref{appendix:discretization}).
See \Cref{fig:tanh_exploration_distance} for a version with plausible momentum in tanh RNNs.
}
\label{fig:non-plaus_exploration_distance}
\end{figure}

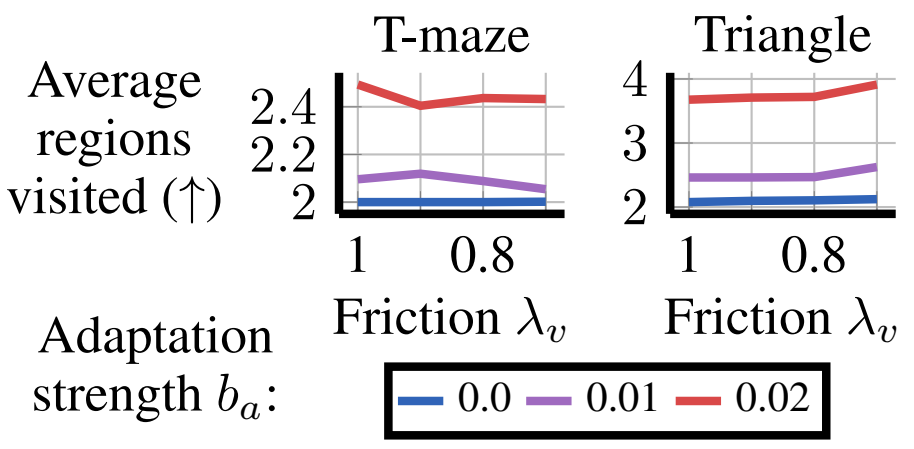
\begin{figure*}[h]
\centering
\begin{tikzpicture}

\pgfplotsset{/pgfplots/group/every plot/.append style = {
    ultra thick,
    legend style={anchor=south west}
}};

\begin{groupplot}[
    group style = {
        group size = 2 by 1,
        horizontal sep = 0.05\textwidth,
        vertical sep = 0.05\textwidth},
    anchor=south west,
    width = 0.22\textwidth,
    height = 0.18\textwidth,
    x dir=reverse,   
    axis x line*=bottom,
    axis y line*=left,
    xlabel={Friction $\friction$},
    xlabel shift={-1mm}, 
    ylabel style={rotate=-90},
    ylabel shift={-3mm},
    xtick={1,0.9,0.8,0.7},
    xticklabels={1,,0.8,},
    grid,
    no marks,
    table/col sep=comma,
    cycle list name=mygradient
    ]

    \nextgroupplot[
        title=T-maze,
        title style={yshift=-2.2mm},
        at={(-100,20)},
        ylabel=\column{Average\\regions\\visited ($\uparrow$)},
        legend style={
            legend columns=3,
            at={(0.2,-1.6)},
            inner sep=1pt,
            font=\footnotesize}]
    \foreach \bval in {0.0,0.01,0.02}{
        \addplot table[x=lambda_v, y={b_a=\bval}] {csv/exploration/exploration_tmaze_region_counts_avg.csv}; 
        \addlegendentryexpanded{ \bval};
    }
        
    \nextgroupplot[
        title=Triangle,
        title style={yshift=-2.9mm},]
    \foreach \bval in {0.0,0.01,0.02}{
        \addplot table[x=lambda_v, y={b_a=\bval}] {csv/exploration/exploration_triangle_region_counts_avg.csv};  
    }

\end{groupplot} 

\node[inner sep=0] at (-0.8,-1.0) {\column{Adaptation\\strength $b_a$:}};

\end{tikzpicture}
\caption{
\textbf{Momentum does not decrease exploration via regions visited in replay from RNNs with non-plausible momentum.}
This is another version of the measurements in \Cref{fig:plaus_region_count_avg}, but with non-plausible momentum (\Cref{appendix:discretization}).
See \Cref{fig:tanh_region_count_avg} for a version with tanh RNNs and plausible momentum.
}
\label{fig:non-plaus_region_count_avg}
\end{figure*}

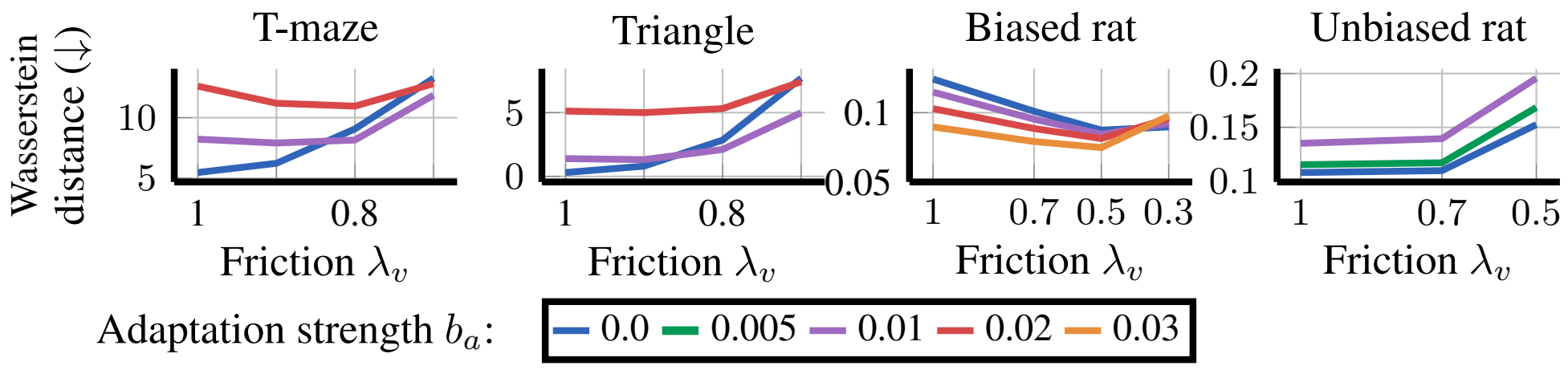
\begin{figure}[h]
\centering
\begin{tikzpicture}

\pgfplotsset{/pgfplots/group/every plot/.append style = {
    ultra thick,
    legend style={
        legend columns=2,
        anchor=south west}
}};

\begin{groupplot}[
    group style = {
        group size = 4 by 1,
        horizontal sep = 0.05\linewidth,
        vertical sep = 0.05\linewidth},
    width = 0.28\linewidth,
    height = 0.18\linewidth,
    title style={yshift=-1.3mm,},
    x dir=reverse,   
    axis x line*=bottom,
    axis y line*=left,
    xlabel={Friction $\friction$},
    xlabel shift={-1mm},
    xticklabel style={font=\small},
    ylabel style={rotate=-90},
    ylabel shift={-3mm},
    yticklabel style={font=\small},
    grid,
    no marks,
    table/col sep=comma,
    cycle list name=mygradient
    ]

    \nextgroupplot[
        title=T-maze,
        ylabel=\column{\small Wasserstein\\distance ($\downarrow$)},
        ylabel style={
            rotate=90,
            yshift=2mm
        },
        xtick={1, 0.9, 0.8, 0.7},
        xticklabels={1, , 0.8, },
        ]
    \foreach \bval in {0.0,0.01,0.02}{
        \addplot table[x=lambda_v, y={b_a=\bval}] {csv/wd/wd_tmaze.csv}; 
    }

    \nextgroupplot[
        title=Triangle,
        title style={yshift=-1mm,},
        xtick={1, 0.9, 0.8, 0.7},
        xticklabels={1, , 0.8, },]
    \foreach \bval in {0.0,0.01,0.02}{
        \addplot table[x=lambda_v, y={b_a=\bval}] {csv/wd/wd_triangle.csv};  
    }

    \nextgroupplot[
        title=Biased rat,
        xtick={1, 0.7, 0.5, 0.3},
        ytick={0.05, 0.1}, 
        yticklabels={0.05, 0.1},
        ymin=0.05,
        legend style={
            at={(-1.3,-1.6)},
            legend columns=5,
            row sep=-3pt,
            inner sep=1pt,
            font=\small}]
    \foreach \bval in {0.0,0.01,0.02,0.03}{
        \addplot table[x=lambda_v, y={b_a=\bval}] {csv/wd/wds_biased.csv};  
        \addlegendentryexpanded{ \bval};
        \ifnum \fpeval{\bval == 0.0} = 1
            \addlegendimage{color=ForestGreen, line width=2pt};
            \addlegendentry{0.005};
        \fi
    }    

    \nextgroupplot[
        title=Unbiased rat,
        xtick={1,0.7,0.5},
        ytick={0.1, 0.15, 0.2}
        ] 
    \addplot[color=C0] table[x=lambda_v, y={b_a=0.0}] {csv/wd/wds_unbiased.csv}; 
    \addplot[color=ForestGreen] table[x=lambda_v, y={b_a=0.005}] {csv/wd/wds_unbiased.csv}; 
    \addplot[color=C1] table[x=lambda_v, y={b_a=0.01}] {csv/wd/wds_unbiased.csv};

\end{groupplot} 

\node[inner sep=0] at (1,-1.25) {\column{Adaptation strength $b_a$:}};

\end{tikzpicture}
\caption{
Wasserstein distance (\Cref{appendix:wasserstein_distance}) of replay trajectories, in RNNs with non-plausible momentum (see \Cref{appendix:discretization}), compared to awake trajectories.
Versions for RNNs with plausible momentum and tanh activations are in Figures \ref{fig:plaus_wd} and \ref{fig:tanh_wd}.
}
\label{fig:non-plaus_wd}
\end{figure}
\section{Results with Tanh Activations}
\label{appendix:tanh}

Throughout all other parts of the text, we have used piecewise-linear (linear or leaky/non-leaky ReLU) activation functions, as defended in \Cref{appendix:hyperparameters}.
For completeness, here we reproduce some key findings of our work with all hyperparameters kept the same, except that RNNs now use tanh activation functions instead of piecewise-linear ones.
We see that momentum temporally compresses replay and boosts exploration, although adaptation does not seem to slow replay in tanh RNNs.

\newcommand{\tanhRtMinTmaze}{-27}
\newcommand{\tanhRtMaxTmaze}{0}
\newcommand{\tanhRtMinTriangle}{-30}
\newcommand{\tanhRtMaxTriangle}{0}
\newcommand{\tanhRtMinBiased}{-62}
\newcommand{\tanhRtMaxBiased}{0}
\pgfplotsset{
    colormap={tanhBwrTmaze}{
        color(\tanhRtMinTmaze)=(blue)
        color(0)=(white)},
    colormap={tanhBwrTriangle}{
        color(\tanhRtMinTriangle)=(blue)
        color(\tanhRtMaxTriangle)=(white)},
    colormap={tanhBwrBiased}{
        color(\tanhRtMinBiased)=(blue)
        color(\tanhRtMaxBiased)=(white)}
}

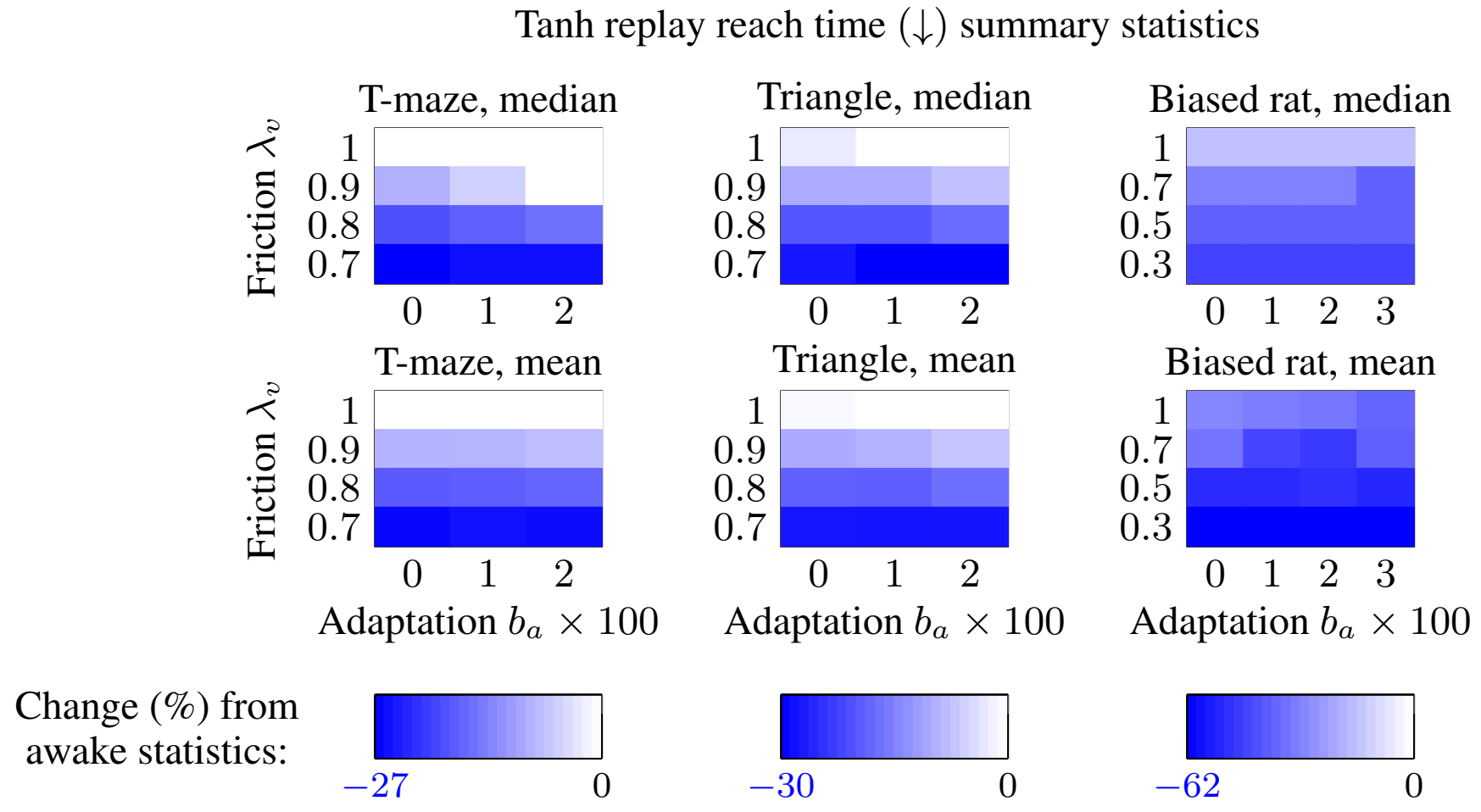
\begin{figure}[h]
\centering
\begin{tikzpicture}

\small

\begin{groupplot}[
    group style = {
        group size = 3 by 2,
        horizontal sep = 0.1\linewidth,
        vertical sep = 0.06\linewidth,
    },
    title style={yshift=-1.7ex,},
    width=0.24\linewidth,
    height=0.2\linewidth,
    xtick={0,1,2},
    xticklabels={$0$,$1$,$2$},
    xlabel style={yshift=0.5ex,},
    ytick={0,1,2,3},
    yticklabels={$1$,$0.9$,$0.8$,$0.7$},
    y dir=reverse,
    enlargelimits={abs=0.5},
    xmin=0, xmax=2,
    ymin=0, ymax=3,      
    ]

    \nextgroupplot[
        title={T-maze, median},
        ylabel=Friction $\friction$,
        point meta min=\tanhRtMinTmaze,
        point meta max=\tanhRtMaxTmaze,    
        colormap name=tanhBwrTmaze,       
    ]
        \addplot [matrix plot*,point meta=explicit] file {csv/rt/plaus_med_rt_heatmap_tmaze.txt};

    \nextgroupplot[
        title={Triangle, median},
        point meta min=\tanhRtMinTriangle,
        point meta max=\tanhRtMaxTriangle,
        colormap name=tanhBwrTriangle,
    ]
        \addplot [matrix plot*,point meta=explicit] file {csv/rt/tanh_med_rt_heatmap_triangle.txt};

    \nextgroupplot[
        title={Biased rat, median},
        point meta min=\tanhRtMinBiased,
        point meta max=\tanhRtMaxBiased,
        colormap name=tanhBwrBiased,
        xmax=3,
        xtick={0,1,2,3},
        xticklabels={$0$, $1$, $2$, $3$},
        yticklabels={$1$, $0.7$, $0.5$, $0.3$}
    ]
        \addplot [matrix plot*,point meta=explicit] file {csv/rt/plaus_tanh_med_rt_heatmap_biased.txt};

    \nextgroupplot[
        title={T-maze, mean},
        point meta min=\tanhRtMinTmaze,
        point meta max=\tanhRtMaxTmaze,
        colormap name=tanhBwrTmaze,  
        ylabel=Friction $\friction$,
        xlabel=Adaptation $b_a \times 100$,
        colorbar sampled, 
        colorbar horizontal,
        colorbar style={
            xtick={\tanhRtMinTmaze, \tanhRtMaxTmaze},
            xticklabels={\textcolor{blue}{$\tanhRtMinTmaze$}, $0$},            
            xticklabel style={font=\footnotesize}
        }        
    ]
        \addplot [matrix plot*,point meta=explicit] file {csv/rt/tanh_avg_rt_heatmap_tmaze.txt};

    \nextgroupplot[
        title={Triangle, mean},
        point meta min=\tanhRtMinTriangle,
        point meta max=\tanhRtMaxTriangle,    
        colormap name=tanhBwrTriangle,    
        xlabel=Adaptation $b_a \times 100$,
        colorbar sampled, 
        colorbar horizontal,
        colorbar style={
            xtick={\tanhRtMinTriangle, \tanhRtMaxTriangle},
            xticklabels={\textcolor{blue}{$\tanhRtMinTriangle$}, $0$},
            xticklabel style={font=\footnotesize}
        }           
    ]
        \addplot [matrix plot*,point meta=explicit] file {csv/rt/tanh_avg_rt_heatmap_triangle.txt};

    \nextgroupplot[
        title={Biased rat, mean},
        point meta min=\tanhRtMinBiased,
        point meta max=\tanhRtMaxBiased,
        colormap name=tanhBwrBiased,
        xmax=3,
        xtick={0,1,2,3},
        xticklabels={$0$, $1$, $2$, $3$},
        yticklabels={$1$, $0.7$, $0.5$, $0.3$},
        xlabel=Adaptation $b_a \times 100$,
        colorbar sampled, 
        colorbar horizontal,
        colorbar style={
            xtick={\tanhRtMinBiased, \tanhRtMaxBiased},
            xticklabels={\textcolor{blue}{$\tanhRtMinBiased$}, $\tanhRtMaxBiased$},
            xticklabel style={font=\footnotesize}
        }        
    ]
        \addplot [matrix plot*,point meta=explicit] file {csv/rt/plaus_tanh_avg_rt_heatmap_biased.txt};

\end{groupplot}

\node[inner sep=0] at (-1.7,-3.5) {\column{Change (\%) from\\awake statistics:}};

\node[inner sep=0] at (4, 2) {Tanh replay reach time $(\downarrow)$ summary statistics};

\end{tikzpicture}
\caption{
\textbf{Momentum temporally compresses replay in tanh RNNs.}
This is another version of \Cref{fig:plaus_reachtime_heatmaps_all}, but with tanh RNNs.
See \Cref{fig:non-plaus_reachtime_heatmaps_all} for a version with non-plausible momentum and ReLU RNNs.
Notably, adaptation does not slow replay notably in tanh RNNs.
}
\label{fig:tanh_reachtime_heatmaps_all}
\end{figure}

\begin{figure}[h!]
\centering
\begin{tikzpicture}

\pgfplotsset{/pgfplots/group/every plot/.append style = {
    ultra thick,
    legend style={
        legend columns=2,
        anchor=south west}
}};

\begin{groupplot}[
    group style = {
        group size = 4 by 1,
        horizontal sep = 0.05\linewidth,
        vertical sep = 0.05\linewidth},
    width = 0.28\linewidth,
    height = 0.18\linewidth,
    title style={yshift=-1.3mm,},
    x dir=reverse,   
    axis x line*=bottom,
    axis y line*=left,
    xlabel={Friction $\friction$},
    xlabel shift={-1mm},
    xticklabel style={font=\small},
    ylabel style={rotate=-90},
    ylabel shift={-3mm},
    yticklabel style={font=\small},
    grid,
    no marks,
    table/col sep=comma,
    cycle list name=mygradient
    ]

    \nextgroupplot[
        title=T-maze,
        ylabel=\column{\small Mean path\\length ($\uparrow$)},
        xtick={1, 0.9, 0.8, 0.7},
        xticklabels={1, , 0.8, },
        ]
    \foreach \bval in {0.0,0.01,0.02}{
        \addplot table[x=lambda_v, y={b_a=\bval}] {csv/exploration/tanh_exploration_tmaze_distances_avg.csv}; 
    }

    \nextgroupplot[
        title=Triangle,
        title style={yshift=-1mm,},
        xtick={1, 0.9, 0.8, 0.7},
        xticklabels={1, , 0.8, },]
    \foreach \bval in {0.0,0.01,0.02}{
        \addplot table[x=lambda_v, y={b_a=\bval}] {csv/exploration/tanh_exploration_triangle_distances_avg.csv};  
    }

    \nextgroupplot[
        title=Biased rat,
        xtick={1, 0.7, 0.5, 0.3},
        legend style={
            at={(-1.3,-1.6)},
            legend columns=5,
            row sep=-3pt,
            inner sep=1pt,
            font=\small}]
    \foreach \bval in {0.0,0.01,0.02,0.03}{
        \addplot table[x=lambda_v, y={b_a=\bval}] {csv/exploration/plaus_tanh_exploration_distance_biased.csv};
        \addlegendentryexpanded{ \bval};
    }    

    \nextgroupplot[
        title=Unbiased rat,
        xtick={1,0.8,0.7},
        ] 
    \foreach \bval in {0.0,0.01,0.02}{
        \addplot table[x=lambda_v, y={b_a=\bval}] {csv/exploration/plaus_tanh_exploration_distance_unbiased.csv};
    }

\end{groupplot} 

\node[inner sep=0] at (1,-1.25) {\column{Adaptation strength $b_a$:}};

\end{tikzpicture}
\caption{
\textbf{Momentum and adaptation both increase exploration via replay path length in tanh RNNs.}
This is another version of \Cref{fig:plaus_exploration_metrics}, but with tanh RNNs.
See \Cref{fig:non-plaus_exploration_distance} for a version with non-plausible momentum in ReLU RNNs.
}
\label{fig:tanh_exploration_distance}
\end{figure}
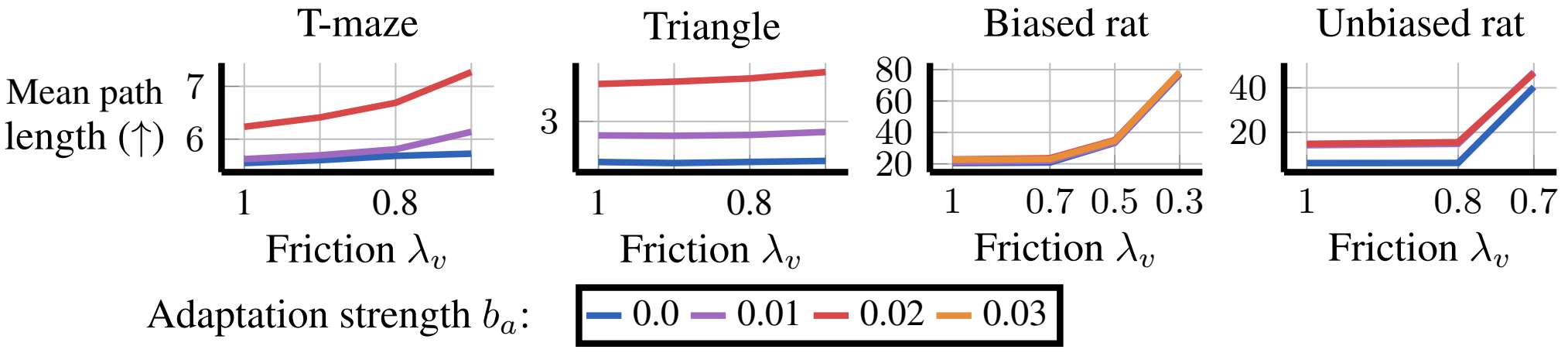

\begin{figure*}[h]
\centering
\begin{tikzpicture}

\pgfplotsset{/pgfplots/group/every plot/.append style = {
    ultra thick,
    legend style={anchor=south west}
}};

\begin{groupplot}[
    group style = {
        group size = 2 by 1,
        horizontal sep = 0.05\textwidth,
        vertical sep = 0.05\textwidth},
    anchor=south west,
    width = 0.22\textwidth,
    height = 0.18\textwidth,
    x dir=reverse,   
    axis x line*=bottom,
    axis y line*=left,
    xlabel={Friction $\friction$},
    xlabel shift={-1mm}, 
    ylabel style={rotate=-90},
    ylabel shift={-3mm},
    xtick={1,0.9,0.8,0.7},
    xticklabels={1,,0.8,},
    grid,
    no marks,
    table/col sep=comma,
    cycle list name=mygradient
    ]

    \nextgroupplot[
        title=T-maze,
        title style={yshift=-2.2mm},
        at={(-100,20)},
        ylabel=\column{Average\\regions\\visited ($\uparrow$)},
        legend style={
            legend columns=3,
            at={(0.2,-1.6)},
            inner sep=1pt,
            font=\footnotesize}]
    \foreach \bval in {0.0,0.01,0.02}{
        \addplot table[x=lambda_v, y={b_a=\bval}] {csv/exploration/tanh_exploration_tmaze_region_counts_avg.csv}; 
        \addlegendentryexpanded{ \bval};
    }
        
    \nextgroupplot[
        title=Triangle,
        title style={yshift=-2.9mm},]
    \foreach \bval in {0.0,0.01,0.02}{
        \addplot table[x=lambda_v, y={b_a=\bval}] {csv/exploration/tanh_exploration_triangle_region_counts_avg.csv};  
    }

\end{groupplot} 

\node[inner sep=0] at (-0.8,-1.0) {\column{Adaptation\\strength $b_a$:}};

\end{tikzpicture}
\caption{
\textbf{Momentum does not decrease exploration via regions visited in replay from tanh RNNs.}
This is another version of the measurements in \Cref{fig:plaus_region_count_avg}, but with tanh RNNs.
See \Cref{fig:non-plaus_region_count_avg} for a version with non-plausible momentum in ReLU RNNs.
}
\label{fig:tanh_region_count_avg}
\end{figure*}
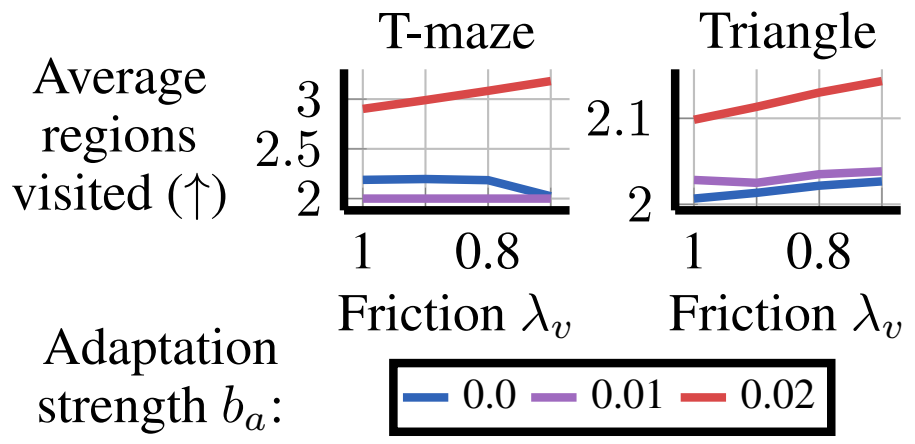

\begin{figure}[h]
\centering
\begin{tikzpicture}

\pgfplotsset{/pgfplots/group/every plot/.append style = {
    ultra thick,
    legend style={
        legend columns=2,
        anchor=south west}
}};

\begin{groupplot}[
    group style = {
        group size = 4 by 1,
        horizontal sep = 0.05\linewidth,
        vertical sep = 0.05\linewidth},
    width = 0.28\linewidth,
    height = 0.18\linewidth,
    title style={yshift=-1.3mm,},
    x dir=reverse,   
    axis x line*=bottom,
    axis y line*=left,
    xlabel={Friction $\friction$},
    xlabel shift={-1mm},
    xticklabel style={font=\small},
    ylabel style={rotate=-90},
    ylabel shift={-3mm},
    yticklabel style={font=\small},
    grid,
    no marks,
    table/col sep=comma,
    cycle list name=mygradient
    ]

    \nextgroupplot[
        title=T-maze,
        ylabel=\column{\small Wasserstein\\distance ($\downarrow$)},
        ylabel style={
            rotate=90,
            yshift=2mm
        },
        xtick={1, 0.9, 0.8, 0.7},
        xticklabels={1, , 0.8, },
        ]
    \foreach \bval in {0.0,0.01,0.02}{
        \addplot table[x=lambda_v, y={b_a=\bval}] {csv/wd/tanh_wd_tmaze.csv}; 
    }

    \nextgroupplot[
        title=Triangle,
        title style={yshift=-1mm,},
        xtick={1, 0.9, 0.8, 0.7},
        xticklabels={1, , 0.8, },]
    \foreach \bval in {0.0,0.01,0.02}{
        \addplot table[x=lambda_v, y={b_a=\bval}] {csv/wd/tanh_wd_triangle.csv};  
    }

    \nextgroupplot[
        title=Biased rat,
        xtick={1, 0.7, 0.5, 0.3},
        ytick={0.08, 0.1, 0.12},
        yticklabels={0.08, 0.1, 0.12},
        legend style={
            at={(-1.3,-1.6)},
            legend columns=5,
            row sep=-3pt,
            inner sep=1pt,
            font=\small}]
    \foreach \bval in {0.0,0.01,0.02,0.03}{
        \addplot table[x=lambda_v, y={b_a=\bval}] {csv/wd/plaus_tanh_wds_biased.csv};
        \addlegendentryexpanded{ \bval};
    }    

    \nextgroupplot[
        title=Unbiased rat,
        xtick={1,0.8,0.7}
        ] 
    \foreach \bval in {0.0,0.01,0.02}{
        \addplot table[x=lambda_v, y={b_a=\bval}] {csv/wd/plaus_tanh_wds_unbiased.csv};
    }

\end{groupplot} 

\node[inner sep=0] at (1,-1.25) {\column{Adaptation strength $b_a$:}};

\end{tikzpicture}
\caption{
Wasserstein distance (\Cref{appendix:wasserstein_distance}) of replay trajectories in tanh RNNs compared to awake trajectories.
Versions for RNNs with ReLU activations and non-plausible momentum are in Figures \ref{fig:plaus_wd} and \ref{fig:non-plaus_wd}.
}
\label{fig:tanh_wd}
\end{figure}


\end{document}